\documentclass{article}
\usepackage[utf8]{inputenc}
\usepackage{arxiv-commands}
\usepackage{soul}
\usepackage{bbm}
\usepackage[numbers]{natbib}

\newcommand{\enayat}[1]{}
\newcommand{\mnote}[1]{}
\newcommand{\cnote}[1]{}
\newcommand{\todo}[1]{}

\newcommand{\remove}[1]{{}}

\title{Differentially Private Generalized Linear Models Revisited}

\author{
\makebox[1.5in]{\hfill Raman Arora\thanks{Department of Computer Science, The Johns Hopkins University,
 \href{mailto:arora@cs.jhu.edu}{arora@cs.jhu.edu} }}
\makebox[1.6in]{\hfill Raef Bassily\thanks{Department of Computer Science \& Engineering and the Translational Data Analytics Institute (TDAI), The Ohio State University,  \href{mailto:bassily.1@osu.edu}{bassily.1@osu.edu} }}
\makebox[1.8in]{\hfill Crist\'obal Guzm\'an
\thanks{Department of Applied Mathematics, University of Twente and Institute for Mathematical and Computational Engineering, Pontificia Universidad Cat\'olica de Chile,
 \href{mailto:c.guzman@utwente.nl}{c.guzman@utwente.nl} }}
\and
\makebox[1.5in]{\hfill Michael Menart \thanks{Department of Computer Science \& Engineering, The Ohio State University,
\href{mailto:menart.2@osu.edu}{menart.2@osu.edu}}}
\makebox[1.5in]{ \hfill Enayat Ullah\thanks{Department of Computer Science, The Johns Hopkins University, \href{mailto:enayat@jhu.edu}{enayat@jhu.edu}}}}

\begin{document}

\date{}

\maketitle

\begin{abstract}
We study the problem of $(\epsilon,\delta)$-differentially private learning of linear predictors with convex losses. We provide results for two subclasses of loss functions. The first case is when the loss is smooth and non-negative but not necessarily Lipschitz (such as the squared loss). For this case, we establish an  upper bound on the excess population risk of $\tilde{O}\left(\frac{\Vert w^*\Vert}{\sqrt{n}} + \min\left\{\frac{\Vert w^* \Vert^2}{(n\epsilon)^{2/3}},\frac{\sqrt{d}\Vert w^*\Vert^2}{n\epsilon}\right\}\right)$, where $n$ is the number of samples, $d$ is the dimension of the problem, and $w^*$ is the minimizer of the population risk. Apart from the dependence on $\Vert w^\ast\Vert$, our bound is essentially tight in all parameters. In particular, we show a lower bound of $\tilde{\Omega}\left(\frac{1}{\sqrt{n}} + {\min\left\{\frac{\Vert w^*\Vert^{4/3}}{(n\epsilon)^{2/3}}, \frac{\sqrt{d}\Vert w^*\Vert}{n\epsilon}\right\}}\right)$. We also revisit the previously studied case of Lipschitz losses \cite{SSTT21}.  For this case, we close the gap in the existing work and show that the optimal rate is (up to log factors) $\Theta\left(\frac{\Vert w^*\Vert}{\sqrt{n}} + \min\left\{\frac{\Vert w^*\Vert}{\sqrt{n\epsilon}},\frac{\sqrt{\text{rank}}\Vert w^*\Vert}{n\epsilon}\right\}\right)$, where $\text{rank}$ is the rank of the design matrix. This improves over existing work in the high privacy regime. Finally, our algorithms involve a private model selection approach that we develop to enable attaining the stated rates without a-priori knowledge of $\Vert w^*\Vert$. 
\end{abstract}
\section{Introduction}
Ensuring privacy of users' data in machine learning algorithms is an important desideratum for multiple domains, such as social and medical sciences. 
Differential privacy (DP) has become the gold standard of privacy-preserving data analysis as it offers formal and quantitative privacy guarantees and enjoys many attractive properties from an algorithmic design perspective \citep{dwork2014algorithmic}. However, for various basic machine learning tasks, the known risk guarantees are potentially sub-optimal and pessimistic.  

In this work, we make progress  towards resolving one of the 
most basic machine learning problems under differential privacy:
learning linear predictors with convex losses.

\subsection{Related Work}
Differentially private machine learning has been thoroughly studied for over a decade. 
In the Lipschitz-convex setting, tight rates are known for both the empirical and population risk \cite{bassily2014private,bassily2019private}. Specifically, it was shown that in the constrained setting, dependence on the dimension in the form of $\Omega\big(\frac{\sqrt{d}}{n\epsilon}\big)$ is unavoidable even for generalized linear models (GLM) (see Section \ref{sec:prelim} for a formal definition).
By contrast, in the unconstrained setting, it has been shown that dimension independent rates are possible for GLMs 
\cite{JainThakurta:2014}. In this setting, assuming prior knowledge of $\mnorm$, the best known rate is  
$O\br{\frac{\mnorm}{\sqrt{n}}+\frac{\mnorm\sqrt{\rank}}{n\epsilon}}$ \cite{SSTT21}, where $\rank$ is the expected rank of the design matrix.
However, without prior knowledge of $\mnorm$, these methods exhibit quadratic dependence on $\mnorm$.
Furthermore, these results crucially rely on the assumption that the loss is Lipschitz to bound the sensitivity. Although gradient clipping has been proposed to remedy this problem \cite{SSTT21, Xiangyi_clipping}, it is known that the solution obtained by clipping may not coincide with the one of the original model. 

Without Lipschitzness, work on differentially private GLMs has largely been limited to linear regression \cite{WGS18, CWZ20}. Here, dimension independent rates have only been obtained  under certain sparsity assumptions. 

More generally, smooth non-negative losses have been studied in the non-private setting by \cite{SST10}, where it was shown such functions can obtain risk guarantees with linear dependence on the minimizer norm (as in the Lipschitz case). This work also established a lower bound of $\Omega\big(\frac{1}{\sqrt{n}}\big)$ on the excess population risk for this class of loss functions. \cite{shamir_2015} additionally establishes a lower bound of 
$\Omega\bc{\min\bc{\frac{\mnorm^2+d}{n},\frac{\mnorm}{\sqrt{n}}}}$ on the excess population risk 
by way of linear regression\footnote{The \cite{SST10} bound assumes $\ybound,H,\xbound=\Omega(1)$. The bounds of \cite{shamir_2015} were originally stated for the constrained setting, but can easily be converted. More details in Appendix \ref{app:nonpriv-lower}.}. 

\subsection{Our Contributions}
\begin{table*}[t]
    \centering
    \resizebox{\textwidth}{!}{
    \begin{tabular}{|c|c|c|c|c|c|c|c|}
 \hhline{~~------}
    \multicolumn{2}{c|}{} & \multicolumn{4}{c|}{ $H$-Smooth, Non-negative} &
    \multicolumn{2}{c|}{$G$-Lipschitz} \\
   \cline{3-8}
      \multicolumn{2}{c|}{} & \multicolumn{2}{c|}{ $ \sqrt{H}\norm{w^*}\norm{\cX} \leq \norm{\cY}$} &  \multicolumn{2}{c|}{ $\sqrt{H}\norm{w^*}\norm{\cX}>\norm{\cY}$} & \multicolumn{2}{c|}{}   \\
    \cline{3-8}
    \multicolumn{2}{c|}{} & $d\leq \br{\frac{\norm{w^*}\sqrt{H}\norm{\cX}n\epsilon}{\norm{\cY}}}^{2/3}$  & $d> \br{\frac{\norm{w^*}\sqrt{H}\norm{\cX}n\epsilon}{\norm{\cY}}}^{2/3} $ & $d\leq \br{n\epsilon}^{2/3}$ & $d>\br{n\epsilon}^{2/3}$  & $\rank\leq n\epsilon$ & $\rank > n\epsilon$ \\
    \hline 
    \hline
    \multirow{5}{*}{{\scriptsize DP}} 
     & UB & 
      $\frac{\sqrt{H}\norm{w^*}\norm{\cX}\norm{\cY}\sqrt{d}}{n\epsilon}$
     & 
    $\frac{\br{\sqrt{H}\norm{w^*}\norm{\cX}}^{4/3}\norm{\cY}^{2/3}}{\br{n\epsilon}^{2/3}}$
     & 
    $\frac{H\norm{w^*}^2\norm{\cX}^2\sqrt{d}}{n\epsilon}$
    &
    $\frac{H\norm{w^*}^2\norm{\cX}^2}{\br{n\epsilon}^{2/3}}$
    &
    $\frac{G\norm{w^*}\norm{\cX}\sqrt{\rank}}{n\epsilon}$ 
    &
      $\frac{G\norm{w^*}\norm{\cX}}{\sqrt{n\epsilon}}$ 
    \\
    & & {\scriptsize \cref{thm:smooth-low-dim}} 
    & 
          {\scriptsize  \cref{thm:smooth-jl}}  
            & 
    {\scriptsize \cref{thm:smooth-low-dim}} & 
    {\scriptsize \cref{thm:smooth-jl}, \ref{thm:smooth-output-perturbation}} &
    {\scriptsize \cite{SSTT21}} &
  {\scriptsize
  \cref{thm:lipschitz-jl},
  \ref{thm:lipschitz-output-perturbation}
   } \\
      \hhline{~-------}
     & 
   LB &  
    Tight
   &  
Tight
   &  
    $\frac{\sqrt{H}\norm{w^*}\norm{\cX}\norm{\cY}\sqrt{d}}{n\epsilon}$
     & 
    $\frac{\br{\sqrt{H}\norm{w^*}\norm{\cX}}^{4/3}\norm{\cY}^{2/3}}{\br{n\epsilon}^{2/3}}$
    &
    Tight
    &
       Tight
    \\
       & &  
      {\scriptsize \cref{thm:smooth-lower-bound}} & 
      {\scriptsize \cref{thm:smooth-lower-bound}} &
       {\scriptsize \cref{thm:smooth-lower-bound}} &
       {\scriptsize \cref{thm:smooth-lower-bound}} &
       {\scriptsize
      \cref{thm:lipschitz_lower}
       } &
       {\scriptsize
      \cref{thm:lipschitz_lower}
      }
       \\
    \hline
      \multirow{4}{*}{{\scriptsize Non-private}} 
      & UB 
      & 
      \multicolumn{4}{c|}{ $\frac{\sqrt{H}\norm{\cX}\norm{\cY}\norm{w^*}}{\sqrt{n}}$} &
      \multicolumn{2}{c|}{ $\frac{G\norm{w^*}\norm{\cX}}{\sqrt{n}}$}\\
     & & 
     \multicolumn{4}{c|}{
     \scriptsize
     \cite{SST10}}  &
    \multicolumn{2}{c|}{
    \scriptsize
    \cite{Nemirovsky:1983}}
    \\
     \hhline{~-------}
     & LB 
        & 
      \multicolumn{4}{c|}{ $\frac{\min\bc{\ybound,\norm{w^*}\xbound}}{\sqrt{n}} + \min\bc{\ybound^2, \frac{H\norm{w^*}^2\xbound^2+d\ybound^2}{n},\frac{\sqrt{H}\norm{w^*}\ybound\xbound}{\sqrt{n}}}$} &
        \multicolumn{2}{c|}{
        Tight
        }\\
     & & 
     \multicolumn{4}{c|}{
     \scriptsize
     \cite{SST10, shamir_2015}} &
    \multicolumn{2}{c|}{ 
    \scriptsize
    \cite{Nemirovsky:1983}
    }\\
    \hline
    \end{tabular}}
        \caption{{ \small Summary of Rates. Parameters: $d$: dimension, $n$: sample size, $H$: smoothness parameter, $w^*$: minimum norm population risk minimizer, $\xbound$: bound on feature vectors, $\ybound$: bound on loss at zero, $G$: Lipschitzness parameter, $\mathsf{rank}$: expected rank of the design matrix, $\epsilon$: privacy parameter ($\delta$ factors omitted). 
        The actual private excess risk bounds are the sum of the expressions shown in the DP rows and their non-private counterparts. 
        Details on non-private lower bounds in Appendix \ref{app:nonpriv-lower}.
        }}
    \label{tab:my_label}
\end{table*}

\paragraph{Smooth nonegative GLMs.} 
Our primary contribution is a new and nearly optimal rate for the problem of differentially private learning of smooth GLMs. In this setting, we focus on characterizing the excess risk in terms of $n,d,\epsilon$ and $\mnorm$. Specifically, we show that it is possible to achieve a rate of 
$\tilde{O}\br{\frac{\norm{w^*}}{\sqrt{n}} + \min\bc{\frac{\norm{w^*}^2}{(n\epsilon)^{2/3}},\frac{\sqrt{d}\mnorm^2}{n\epsilon}}}$ on the excess population risk. 
Our new rates exhibit an interesting low/high dimensional transition at $d\approx (\mnorm n\epsilon)^{2/3}$.  
First, in the low dimensional regime, we develop a novel analysis of noisy gradient decent (GD) inspired by techniques from \cite{SST10}. 
In particular, we show that Noisy GD gives an improved rate for non-negative smooth functions (not necessarily GLMs). This is based on an average stability analysis of Noisy GD. As we elaborate in Section
\ref{sec:smooth_upper},
a straightforward application of uniform stability leads to sub-optimal bounds and hence a new analysis is required.
We note in passing that this upper bound works for (unconstrained) DP-SCO with smooth (non-Lipschitz) losses, which is of independent interest.
For the high dimensional regime, we perform random projections of the data (specifically, the Johnson-Lindenstrauss transform) for dimensionality reduction, roughly reducing the problem to its low dimensional counterpart. 
We also develop a lower bound for the excess risk under DP of $\tilde{\Omega}\br{{\min\bc{\frac{\norm{w^*}^{4/3}}{(n\epsilon)^{2/3}}, \frac{\sqrt{d}\mnorm}{n\epsilon}}}}$. 
We note that non-privately a lower bound of $\tilde{\Omega}\br{\frac{1}{\sqrt{n}}+\min\bc{\frac{d+\mnorm^2}{n}, \frac{\mnorm}{\sqrt{n}}}}$ is known on the excess population risk \cite{SST10,shamir_2015}.
We note that these private and non-private lower bounds imply that our bound is optimal up to factors of $\mnorm$ (see Table \ref{tab:my_label}). 

\paragraph{Lipschitz GLMs.} 
For the Lipschitz case, we close a subtle but important gap in existing rates. 
In this setting, it has been shown that one can characterize the excess risk in terms of the expected rank of the design matrix, $\rank$, instead of $d$ \cite{SSTT21}.
In this setting, the best known rate was 
$\tilde{O}\br{\frac{\norm{w^*}}{\sqrt{n}} + \frac{\sqrt{\rank}\mnorm}{n\epsilon}}$. 
We show an improved rate of 
$\tilde{O}\br{\frac{\norm{w^*}}{\sqrt{n}} + \min\bc{\frac{\mnorm}{\sqrt{n\epsilon}},\frac{\sqrt{\rank}\mnorm}{n\epsilon}}}$.
This improves in the high privacy regime where $\epsilon \leq \frac{\rank}{n}$. In fact, the upper bound $O\br{\frac{\mnorm}{\sqrt{n\epsilon}}}$ for this rate can be obtained with only minor adjustments to the regularization method of  \cite{JainThakurta:2014}. Our second contribution in this setting is extending the lower bound of \cite{SSTT21} to hold for all values of $\mnorm > 0$ and $\rank \in [n]$.
This is in contrast to the original lower bound which only holds for problem instances where $\mnorm^2 = \rank$ and $\rank \in [n\epsilon]$.

\paragraph{Model selection.}
As part of our methods, we develop a differentially private model selection approach which eliminates the need for a-priori knowledge of $\mnorm$. Although such methods are well established in the non-private case, (see e.g.~\cite{shalev2014understanding}), in the private case no such methods have been established. Our method, as in the non-private case, performs a grid search over estimates of $\mnorm$ and picks the best model based on the loss. However, in the private setting we must account for the fact the the loss evaluation must be privatized. This is non-trivial in the non-Lipschitz smooth case as the loss at a point $w$ may grow quadratically with $\norm{w}$. 

\paragraph{Lower bounds for Non-Euclidean DP-SCO.}
Our lower bound construction generalizes to Non-Euclidean $\ell_p/\ell_q$ variants of DP-SCO with Lipschitz convex losses \cite{bassily2021non}. 
Herein, we assume that the loss function is $G_q$-Lipschitz with respect to $\ell_q$ norm, and radius of the constraint set is bounded in $\ell_p$ norm by $B_p$.
For this setting, we give a lower bound of $\Omega\br{G_qB_p\min\br{\frac{1}{\br{n\epsilon}^{1/p}}, \frac{d^{(p-1)/p}}{n\epsilon}}}$ on excess empirical/population risk of any (potentially unconstrained) $(\epsilon,\delta)$-DP algorithm; see Corollary \ref{corr:lipschitz_dp_scpo_lower_lp} in Appendix \ref{sec:lower-bound-lp} for a formal statement and proof.
For $p=\infty$ and $p\geq 2, d\leq n\epsilon$, this matches the best known upper bounds in \cite{bassily2021non}.

\paragraph{Non-private settings.}
As by-products, we give the following new results for the non-private setting. For details on the parameters used below we refer to Table \ref{tab:my_label}.
\begin{CompactEnumerate}
    \item We show that gradient descent, when run on convex non-negative $\tilde H$ smooth functions (not necessarily GLMs), it achieves the optimal rate of $O\big(\frac{\sqrt{\tilde H}\norm{w^*}\ybound}{\sqrt{n}}\big)$ (see Corollary \ref{cor:smooth-low-dim-non-priv}). This is done via an average-stability analysis of gradient descent.
    This result is interesting as it also shows GD only needs $n$ iterations, which is known not to work for non-smooth SCO   \cite{bassily2020stability,Amir:2021never,Amir:2021sgd}.
    \item In Section \ref{sec:appendix_boosting}, we give a procedure to boost the confidence of algorithms for risk minimization with convex non-negative $\tilde H$ smooth functions (not necessarily GLMs). The standard boosting analysis based on Hoeffding's inequality does not 
    give a bound with a linear dependence on the parameters $(\norm{w^*},\xbound,\ybound)$, 
    and hence a tighter analysis is required. 
\end{CompactEnumerate}

\subsection{Techniques}

\paragraph{Upper bounds.} 
We give two algorithms for both the smooth and Lispschitz cases. The first method is simple and has two main steps. First, optimize the regularized empirical risk over the constraint set $\bc{w:\norm{w}\leq B}$ for some $B\geq \norm{w^*}$. Then 
output a perturbation of the regularized minimizer with Gaussian noise 
(which is not requried to be in the constraint set). This method is akin to that of \cite{JainThakurta:2014} with the modification that the regularized minimizer is constrained to a ball. We elaborate on this key difference shortly. 

The second method is based on dimensionality reduction.
We use smaller dimensional data-oblivious embeddings of the feature vectors.
A linear JL transform suffices to give embeddings with the required properties. 
We then run a constrained DP-SCO method (Noisy GD) in the embedded space, and use the transpose of the JL transform to get a $d$ dimensional model. 
In this method, the embedding dimension required is roughly the threshold on dimension at which the rates switch from dimension dependent to independent bounds. 
We also remark that \cite{NUZ20} applied a similar technique to provide dimension independent classification error guarantees for privately learning support vector machines under hard margin conditions. 

We note that a crucial part in all of these methods is the use of constrained optimization as a subroutine, where the constraint set is a ball of radius $\mnorm$. This is in stark contrast to the Lipschitz case where existing methods such as those presented by \cite{JainThakurta:2014,SSTT21} rely on the fact that projection is not required. In the smooth case however, constrained optimization helps ensure that the norm of the gradient is roughly bounded by the diameter of the constraint set.
We note that in the high dimensional regime, the property that the \emph{final} output of the algorithm can have large norm is still crucial to the success of our algorithms.

\paragraph{Lower bounds.} 
For our lower bounds in the smooth case we rely on the connection between stability and privacy. Specifically, we will utilize a lemma from \citep{CH12} which bounds the accuracy of one-dimensional differentially private algorithms. We then combine this with packing arguments to obtain stronger lower bounds for high dimensional problems. For the Lipschitz case, we adapt the method of \cite{SSTT21}. 
\section{Preliminaries} \label{sec:prelim}
In the following we detail  several concepts needed for the presentation of this paper.
\paragraph{Risk minimization.}
Let $\cX \subseteq \bbR^d$ be the domain of features,
and $\cY \subseteq \bbR$ be the domain of responses. A linear predictor is any $w\in\bbR^d$.
Let $\ell:\re^d \times (\cX\times\cY) \to \re$ be a loss function. 
Given some unknown distribution $\cD$ over $(\cX\times\cY)$, we define the population loss $L(w;\cD) = \ex{(x,y)\sim\cD}{\ell(w;(x,y))}$. Given some dataset $S\in(\cX\times\cY)^n$ drawn i.i.d. from $\cD$, the objective is to obtain $\out\in\re^d$ which minimizes the excess population risk,
$L(\out;\cD) - \min\limits_{w\in\re^d}\bc{L(w;\cD)}$.
Given a population risk minimization problem, we will denote $w^{\ast}$ to be a \emph{minimum norm} solution to this problem.
We define the empirical risk as $\eloss(w;S)=\frac{1}{n}\sum\limits_{(x,y)\in S}\ell(w;(x,y))$.
We define the following quantities for notational convenience:
$\excessrisk(w) = \ploss(w;\cD)-\ploss(w^*;\cD)$, 
$\excessermrisk(w) = \eloss(w;S)-\eloss(w^*;S)$, 
and 
$\generr(w) = \ploss(w;\cD)-\eloss(w;S)$. We define $\ball{B}$ to be the Euclidean ball of radius $B$ on $\bbR^d$.

\paragraph{Generalized linear models.}
We will more specifically be interested in the problem of learning generalized linear models, where there exists some function $\phi:\re \times \cY \rightarrow \re$ such that the loss function can be written as 
$\ell(w;(x,y)) = \phi(\ip{w}{x},y)$. 
We define parameter bounds $\norm{\cX}=\max_{x\in\cX}{\norm{x}}$ and $\norm{\cY}^2=\max_{y\in\cY}{|\phi(0,y)|}$. Note that for many common 
loss functions, the latter condition is the moral equivalent of assuming labels bounded by $\ybound$. 
For ease of notation, we write $\phi(\ip{w}{x},y)$ as $\phi_y(\ip{w}{x})$ and denote function $\phi_y: z\mapsto
 \phi(z,y)$.
 We say that the loss function is $G$-Lipschitz GLM if  all $y \in \cY$, $\phi_y:\bbR \rightarrow \bbR$ is $G$-Lipschitz. We similarly define $H$-smooth GLM.

\paragraph{Differential privacy~\cite{DKMMN06}.}  
We restrict our investigation to the class of algorithms which minimize the excess population risk under the constraint of differential privacy. 
A randomized algorithm $\cA$ is said to be $(\epsilon,\delta)$ differentially private (i.e., $(\epsilon,\delta)$-DP) if for any pair of datasets $S$ and $S'$ differing in one point and any event $\mathcal{E}$ in the range of $\cA$ it holds that 
\[
\mathbb{P}[\cA(S)\in\mathcal{E}] \leq e^{\epsilon}\mathbb{P}[\cA(S')\in \mathcal{E}] + \delta.
\]

For our lower bounds, we will make use of the following Lemma from \cite{CH12}.
\begin{lemma} \label{lem:ch}
Let $\cZ$ be a data domain and let $S$ and $S'$ be two datasets each in $\cZ^n$ that differ in at most $\Delta$ entries, and let $\cA:\cZ^n\to \re$ be any $(\epsilon,\delta)$-DP algorithm. For all $\tau\in\re$, if $\Delta \leq \frac{\log{1/2\gamma}}{\epsilon}$ and $\delta \leq \frac{1}{16}(1-e^{-\epsilon})$, then
$  \ex{}{|\cA(S) - \tau| + |\cA(S')-\tau'|} \geq \frac{1}{4}|\tau-\tau'|.$
\end{lemma}
Finally, we introduce the Johnson-Lindenstrauss (JL) transform to perform random projections. 
\begin{definition}[$(\alpha,\beta)$-JL property] A distribution over matrices $\mathbb{R}^{k\times d}$ satisfies $(\alpha,\beta)$-JL property if for any 
$u,v \in\bbR^d$, $\mathbb{P}\left[\abs{\ip{\Phi u}{\Phi v} - \ip{u}{v}} > \alpha\norm{u}\norm{v}\right]\leq \beta.$

\label{def:jl}
\end{definition}
It is well known that several such ``data-oblivious'' (i.e. independent of $u$,$v$) distributions exist with $k = O\br{\frac{\log{1/\beta}}{\alpha^2}}$ \cite{Nelson2011SketchingAS}.
We note that the JL property is typically described as approximation of norms (or distances), but it is easy to deduce the above dot product preservation property from it; for completeness we give this as Lemma \ref{lem:jl}.
Finally, we use $\Phi \cD$ to denote the push-forward measure of the distribution 
$\cD$ under the map $\Phi: (x,y)\mapsto (\Phi x,y)$. Similarly, given a data set $S=\bc{(x_i,y_i)}_{i}$, we define $\Phi S:=\bc{\br{\Phi x_i,y_i}_i}$
\section{Smooth Non-negative GLMs} \label{sec:smooth_glm}
For smooth non-negative GLMs, we present new upper and lower bounds on the excess risk. 
For our upper bounds, we here assume that the algorithm is given access to some upper bound on $\mnorm$, that we denote by $B$. 
We later show in Section \ref{sec:adapt} how to obtain such a rate without prior knowledge of $\mnorm$. We also emphasize that the privacy of these algorithms holds regardless of whether or not $B \geq \mnorm$. 

\subsection{Upper Bounds} \label{sec:smooth_upper}
Before presenting our algorithms, we highlight some key ideas underlying all our methods.
A crucial property of non-negative smooth loss functions which allows one to bound sensitivity, and thus ensure privacy, is the self-bounding property (e.g.~Sec.~12.1.3 in \cite{shalev2014understanding}), which states that for an $\tilde H$-smooth non-negative function $f$ and $u\in\operatorname{dom(f)}$, $\norm{\nabla f(u)} \leq \sqrt{4\tilde H f(u)}$. 
 
This property implies that the gradient grows at most linearly with $\norm{w}$. 
More precisely, we have the following.
\begin{lemma} \label{lem:smooth_lip}
Let $\ell$ be an $H$-smooth non-negative GLM. Then for any $w\in\ball{B}$ and $(x,y)\in(\cX\times\cY)$ we have 
$\norm{\nabla\ell(w,(x,y))} \leq 2\norm{\cY}\sqrt{H}\xbound + 2HB\xbound^2$.
\end{lemma} 
In order to leverage this property, all our algorithms in this setting utilize constrained optimization as a subroutine, where the constraint set is $\ball{B}$ and the Lipschitz constant is $G=2\norm{\cY}\sqrt{H}\xbound + 2HB\xbound^2$. This, in conjunction with the self-bounding property ensures reasonable bounds on sensitivity. In turn, this allows us to ensure privacy 
without excessive levels of 
noise.

Finally, we note that our upper bounds for smooth GLMs distinguish low and high dimensional regimes, 
transitioning at $d=\min\Big(\br{\frac{B\sqrt{H}\xbound}{\ybound}}^{\frac{2}{3}}\!\!,\,1\Big)(n\epsilon)^{\frac{2}{3}}$.

\subsubsection{
Low~dimensional regime}

We start with the low dimensional setting where we use techniques developed for constrained DP-SCO for Lipschitz losses (not necessarily GLMs).
Existing private algorithms for DP-SCO (e.g., \cite{bassily2019private,FKT:2020}) lead to excess risk bounds that scale with $\frac{HB^2}{\sqrt{n}}$. 
On the other hand, the optimal non-private rate \cite{SST10} scales  with $\frac{\sqrt{H}B}{\sqrt{n}}$, which may indicate that the private rate implied by the known methods is sub-optimal. 
We show that this gap can be closed by a novel analysis of private GD. 

A standard proof of excess risk of (noisy) gradient descent for smooth convex functions is based on uniform stability \cite{HRS15,bassily2019private}. 
However, this still leads to sub-optimal rates. 
Hence we turn to an average stability based analysis of GD,
yielding 
the following result.

\begin{theorem}
\label{thm:smooth-low-dim}
Let $\ell$ be a non-negative, convex, $\tilde H$-smooth loss function, bounded at zero by $\norm{\cY}^2$. 
Let $B, n > 0$, $n_0 = \frac{\tilde HB^2}{\ybound^2}$,
$G=2\norm{\cY}\sqrt{\tilde H} + 2\tilde HB$. Then, for any $\epsilon, \delta>0$, Algorithm~\ref{alg:noisySGD} invoked with $\cW=\ball{B}$, $T=n$, $\sigma^2 = \frac{8G^2T\log{1/\delta}}{n^2\epsilon^2}$, 
$\eta = \min\Big(\frac{B}{\sqrt{T}\max\big(\sqrt{\tilde H}\ybound, \sigma\sqrt{d}\big)}, \frac{1}{4\tilde H}\Big)$ is $(\epsilon, \delta)$-differentially private. 
Further, given a dataset $S$ of $n\geq n_0$ i.i.d samples from an unknown distribution $\cD$, the excess risk of output of Algorithm \ref{alg:noisySGD} is bounded as,
\begin{align*}
    \mathbb{E}[\excessrisk(\hat w)] \leq O\Bigg(
    \frac{\sqrt{\tilde H}B\ybound}{\sqrt{n}} + 
    \frac{GB\sqrt{d \log{1/\delta}}}{n\epsilon} \Bigg).
\end{align*}
\end{theorem}
We note that since $H$-smooth GLMs satisfy the Theorem condition with parameter $\tilde H\leq H\xbound^2$, we obtain results for GLMs as a direct corollary. 

\paragraph{Non-private risk bound:} As a corollary (see Corollary \ref{cor:smooth-low-dim-non-priv} in Appendix \ref{sec:app_smooth_upperbounds_lowd}), with no privacy constraint, the above result (setting $\epsilon \rightarrow \infty$ and  $\delta = 1$) shows that gradient descent achieves the optimal excess risk bound, previously shown to be
achievable by regularized ERM and one-pass SGD \cite{SST10}.
The lower bound, $n_0$, simply means that the trivial solution ``zero" has larger excess risk. 

\remove{
\begin{corollary} \label{cor:smooth_low_dim_glm}
Let $n_0 = \frac{H\xbound^2B^2}{\ybound^2}$,
$\eta = \min\Big(\frac{B}{\sqrt{T}\max\br{\sqrt{ H}\xbound\ybound, \sigma\sqrt{d}}}, \frac{1}{4H \xbound}\Big)$,  $\cW = \ball{B}$, $\sigma^2 = \frac{8G^2T\log{1/\delta}}{n^2\epsilon^2}$  and 
$T=n$.
Let $\ell$ be a non-negative convex $H$ smooth GLM, bounded at zero by $\norm{\cY}^2$. 
Algorithm \ref{alg:noisySGD} satisfies $(\epsilon, \delta)$-differential privacy.
Given a dataset $S\sim\cD^n$, $n\geq n_0$, then
the excess risk of output of Algorithm \ref{alg:noisySGD} is bounded as,
\begin{align*}
    &\ex{}{\excessrisk(\out)} 
    = O\Bigg(
    \frac{\sqrt{H}\xbound B\ybound}{\sqrt{n}} \\
    &+ \frac{\br{\sqrt{H}\xbound\ybound + H \xbound^2B^2}B\sqrt{d \log{1/\delta}}}{n\epsilon}\Bigg).
\end{align*}
\end{corollary}
}

\paragraph{Proof sketch of Theorem \ref{thm:smooth-low-dim}.}
The privacy proof simply follows from \cite{bassily2014private} since the loss function is $G$-Lipschitz in the constraint set.
For utility, we first introduce two concepts used in the proof. 
Let $S$ be a dataset of $n$ i.i.d samples $\bc{(x_i,y_i)}_{i=1}^n$,  $S^{(i)}$ be the dataset where the $i$-th data point is replaced by an i.i.d. point $(x',y')$.
Let $\cA$ be an algorithm which takes a dataset as input and outputs $\cA(S)$. The \textbf{average argument stability} of $\cA$, denoted as $\uas$ 
is defined as 
\begin{align*}
    \uas^2 = \mathbb{E}_{S,i, (x',y')}\|\cA(S)-\cA(S^{(i)})\|^2.
\end{align*}

The  \textbf{average regret} of gradient descent (Algorithm \ref{alg:noisySGD}) with iterates $\bc{w_t}_{t=1}^T$ is
\begin{align*}
    \textstyle\averageregret(\cA;w^*) = \frac{1}{T}\sum_{j=1}^T \mathbb{E}[\eloss(w_j;S) - \eloss(w^*;S)].
\end{align*}

The key arguments are as follows: we first bound the generalization error, or on-average stability, 
in terms of average argument stability and excess empirical risk (Lemma \ref{lem:gen-gap-uas}). We then bound average argument stability in terms of average regret (Lemma \ref{lem:gd-average-stability}).
Finally, in Lemma \ref{lem:noisy-sgd-erm}, we provide bounds on excess empirical risk and average regret of noisy gradient descent. 
Substituting these in the excess risk decomposition gives the claimed bound. The full proof with the above lemmas is deferred to Appendix \ref{sec:app_smooth_upperbounds_lowd}.

\remove{
These key lemmas are stated below. 
\begin{lemma}
\label{lem:gen-gap-uas}
Let $n\geq \frac{\tilde H B^2}{\ybound^2}$.
Let $\ell$ be a non-negative $\tilde H$ smooth convex loss function.
Given a dataset $S$ be of $n$ i.i.d. samples from an unknown distribution $\cD$, then
 
\begin{align*}
    \mathbb{E}[\generr(\cA(S);\cD,S)] \leq   \frac{\sqrt{\tilde H}B}{\sqrt{n}\norm{\cY}}
    \mathbb{E}[\excessermrisk(\cA(S))] \\+
    \frac{2\sqrt{\tilde H n}\ybound}{B}\uas^2
    + \frac{\sqrt{\tilde H}B\ybound}{\sqrt{n}}.
\end{align*}    
\end{lemma}

\begin{lemma}
\label{lem:gd-average-stability}
The average argument stability for noisy GD (Algorithm \ref{alg:noisySGD}) run for $T$ iterations with step size $\eta \leq \frac{4}{\tilde H}$ is bounded as 
  $$\uas^2 \leq \frac{8\tilde H \eta^2 T}{n} \averageregret(\cA;w^*) + \frac{8\tilde H \eta^2 T\ybound^2}{n}.$$
\end{lemma}

\begin{lemma}
\label{lem:noisy-sgd-erm}
Let $n\geq \frac{\tilde HB^2}{\ybound^2}$, $\eta = \min\Big(\frac{B}{\sqrt{T}\max\big(\sqrt{\tilde H}\ybound, \sigma\sqrt{d}\big)}, \frac{1}{4\tilde H}\Big)$,  $\sigma^2 = \frac{8G^2T\log{1/\delta}}{n^2\epsilon^2}$ and 
$T=n$.
We have
\begin{align*}
    &\mathbb{E}[\excessermrisk(\cA(S))]
    \leq \averageregret(\cA;w^*) \\&= O\br{\frac{\sqrt{H}B\ybound}{\sqrt{n}} + \frac{\sqrt{d}G\log{1/\delta}}{n\epsilon}}.
\end{align*}
\end{lemma}
}

\begin{algorithm}[h]
\caption{Noisy GD}
\label{alg:noisySGD}
\begin{algorithmic}[1]
\REQUIRE Dataset $S$, 
loss function $\ell$, constraint set $\cW$, steps $T$, learning rate $\eta$, noise scale $\sigma^2$ 
\STATE $w_0 = 0$
\FOR{$t=1$ to $T-1$}
\STATE $\xi \sim \cN(0,\sigma^2\bbI)$
\STATE $w_{t+1} = \Pi_{\cW}\br{w_t - \eta \br{\eloss(w_t;S)+\xi}}$
\newline where $\Pi_{\cW}$ is the Euclidean projection on to $\cW$
\ENDFOR
\ENSURE{$\hat w = \frac{1}{T}\sum_{t=1}^Tw_j$}
\end{algorithmic}
\end{algorithm}

\subsubsection{High~dimensional regime}
\begin{algorithm}[t]
\caption{JL Method}
\label{alg:jl-constrained-dp-erm}
\begin{algorithmic}[1]
\REQUIRE Dataset $S=\bc{(x_1,y_1),...,(x_n,y_n)}$, loss function $\ell$,  $k$, $B, \eta, T$, $\sigma^2$
\STATE Sample JL matrix $\Phi \in \re^{k \times d}$
\STATE $\tilde S = \bc{\br{\Phi x_1,y_1},\br{\Phi x_2,y_2}, \ldots, \br{\Phi x_n,y_n}}$
\STATE $\tilde w = \operatorname{NoisyGD}(\tilde S, \ell, \ball{2B}, T, \eta, \sigma^2)$
\ENSURE{$\hat w = \Phi^\top \tilde w$}
\end{algorithmic}
\end{algorithm}

In the high dimensional setting,
we present two techniques.

\paragraph{JL method.}
In the JL method, Algorithm \ref{alg:jl-constrained-dp-erm}, we use a data-oblivious JL map $\Phi$ to embed all feature vectors in dataset $S$ to $k<d$ dimensions.
Let dataset 
$\tilde S = \bc{\br{\Phi x_i,y_i}}_{i=1}^n$.
We then run projected Noisy GD method (Algorithm \ref{alg:noisySGD})
on the loss with dataset $\tilde S$ and the diameter of the constraint set as $2B$. Finally, we map the returned output $\tilde w$ back to $d$ dimensions using $\Phi^\top$ to get $\hat w = \Phi^\top \tilde w$. We note that no projection is performed on $\out$ and thus the output may have large norm due to re-scaling induced by $\Phi^{\top}$.
\begin{theorem}
\label{thm:smooth-jl}
Let $k = O\br{\frac{B\sqrt{H}\norm{\cX}\log {2n/\delta} n\epsilon}{\norm{\cY}\norm{\cX}+ \sqrt{H}B\norm{\cX}^2}}^{2/3}$,
$\cW=\ball{B}$, $T=n$, $\sigma^2 = \frac{8G^2T\log{1/\delta}}{n^2\epsilon^2}$, 
$\eta = \min\Big(\frac{B}{\sqrt{T}\max\big(\sqrt{H}\xbound\ybound, \sigma\sqrt{d}\big)}, \frac{1}{4H\xbound^2}\Big)$
and
$n_0 = \frac{HB^2\norm{\cX}^2}{\norm{\cY}^2}$.
Algorithm \ref{alg:jl-constrained-dp-erm} satisfies $(\epsilon,\delta)$-differential privacy. 
Given a dataset $S$ of $n\geq n_0$ i.i.d samples, 
of the output $\hat w$ is bounded by
\begin{align*}
\mathbb{E}[\excessrisk(\hat w)] \leq
        \tilde O\Bigg(\frac{\sqrt{H}B\norm{\cX}\norm{\cY}}{\sqrt{n}}
      +\frac{\br{\sqrt{H}B\norm{\cX}\sqrt{\norm{\cY}}}^{\frac{4}{3}}+ \br{\sqrt{H}B\norm{\cX}}^2}{(n\epsilon)^{\frac{2}{3}}}\Bigg).
\end{align*}
\end{theorem}

\paragraph{Proof sketch of Theorem \ref{thm:smooth-jl}.}
From the JL property, with $k=O(\log{n/\delta}/\alpha^2)$, w.h.p. norms of all feature vectors and $w^*$, as well as inner products between are preserved upto an $\alpha$ tolerance (see Definition \ref{def:jl}). 
The preservation of norms of feature vectors implies the gradient norms are preserved, and thus privacy guarantee of sub-routine, Algorithm \ref{alg:noisySGD}, suffices to establish DP. 
For the utility proof,
from our analysis of Noisy GD 
(Theorem \ref{thm:smooth-low-dim}) and using the JL property, the excess risk of $\hat w$ under $\cD $ w.r.t. the risk of $\Phi w^*$ under $\Phi \cD$
is bounded as,
\begin{align*}
    \mathbb{E}[L(\hat w;\cD) &- L(\Phi w^*;\Phi\cD)] \leq O\Bigg(
    \frac{\sqrt{H}\xbound B\ybound}{\sqrt{n}}
    + 
    \frac{\br{\sqrt{ H}\xbound\ybound +  H\xbound^2 B^2}B\sqrt{k \log{1/\delta}}}{n\epsilon}\Bigg).
\end{align*}
From smoothness and JL property, the, ``JL error'' is:
\begin{align*}
    \mathbb{E}[L(\Phi w^*; \Phi \cD)]- L(w^*;  \cD)] \leq \tilde O\left(\frac{HB^2\xbound^2}{k}\right).
\end{align*}
The above is optimized for the value of $k$ prescribed in Theorem \ref{thm:smooth-jl}, substituting which gives the claimed bound.

\begin{algorithm}[t]
\caption{Regularized Output Perturbation}
\label{alg:constrained-reg-erm-output-perturbation}
\begin{algorithmic}[1]
\REQUIRE Dataset $S=\bc{(x_1,y_1),...,(x_n,y_n)}$, loss function $\ell$, $\lambda, B, \sigma^2$
\STATE $\tilde w = \argmin\limits_{w\in\ball{B}}\bc{L(w;S) + \frac{\lambda}{2}\norm{w}^2}$
\STATE $\xi \sim \cN(0,\sigma^2\bI_d)$
\ENSURE{$\hat w = \tilde w + \xi$}
\end{algorithmic}
\end{algorithm}
\paragraph{Constrained regularized ERM + output perturbation.}
Our second technique is \emph{constrained} regularized ERM with output perturbation (Algorithm \ref{alg:constrained-reg-erm-output-perturbation}).
A similar technique for the Lipschitz case was seen in \cite{JainThakurta:2014}, however we note that the addition of the constraint set $\ball{B}$ is crucial in bounding the sensitivity in the smooth case.
\begin{theorem}
\label{thm:smooth-output-perturbation}
 Let $n_0=\frac{HB^2\xbound^2}{\ybound^2}$.
Then Algorithm \ref{alg:constrained-reg-erm-output-perturbation} run with 
$\sigma^2 = O\br{\frac{\br{\norm{\cY}^2+H^2B^2\norm{\cX}^4}\log{1/\delta}}{\lambda^2n^2\epsilon^2}}$ and $\lambda = \br{\frac{\br{\norm{\cY}+HB\norm{\cX}^2}\sqrt{H}\norm{\cX}}{Bn\epsilon}}^{2/3}\br{\log{1/\delta}}^{1/3}$
satisfies $(\epsilon,\delta)$-differential privacy. Given a dataset $S$ of $n\geq n_0$ i.i.d samples, the excess risk of its output $\hat w$ is bounded as
\begin{align*}
    \mathbb{E}[\excessrisk(\hat w)]
    &\leq  \tilde O\Bigg(\frac{ \sqrt{H}B\norm{\cX}\norm{\cY} + \ybound^2}{\sqrt{n}} 
    +\frac{\br{\sqrt{H}B\norm{\cX}}^{4/3}\norm{\cY}^{2/3}+ \br{\sqrt{H}B\norm{\cX}}^2}{(n\epsilon)^{2/3}}\Bigg).
\end{align*}

\end{theorem}

We note that we can use the same technique in the low dimensional setting too, yielding a rate of $\frac{ \sqrt{H}B\norm{\cX}\norm{\cY} + \ybound^2}{\sqrt{n}} + \frac{GB\sqrt{d}}{n\epsilon}$.
However, in contrast to Theorem  \ref{thm:smooth-low-dim} and \ref{thm:smooth-jl}, these results have an additional $\frac{\ybound^2}{\sqrt{n}}$ term (in both regimes). Thus, in the regime when $\ybound\leq \sqrt{H}B\xbound$, the two upper bounds are of the same order.

\paragraph{Proof sketch of \cref{{thm:smooth-output-perturbation}}.}
The privacy proof follows the Gaussian mechanism guarantee together with the fact that the $\ell_2$-sensitivity of constrained regularized ERM is $O\br{\frac{G}{\lambda n}}$ \cite{Bousquet:2002}. For utility, we use the Rademacher complexity based result of \cite{SST10} to bound the generalization error of $\tilde w$. The other term, error from noise, $\mathbb{E}[L(\hat w;\cD)-L(\tilde w;\cD)] \leq O(H\sigma^2\xbound^2)$ from smoothness of GLM. Combining these two and optimizing for $\lambda$ gives the claimed result.

\subsection{Lower Bounds} \label{sec:smooth_lower}
The proof technique for our lower bound relies on the connection between differential privacy and sensitivity shown in \cite{CH12}. 
The underlying mechanisms behind the proof is also similar in nature to the method seen in~\cite{steinke2017tight}. 
We note that although we present the following theorem for empirical risk, a reduction found in \cite{bassily2019private} implies the following result holds for population risk as well (up to log factors). 
\begin{theorem}
\label{thm:smooth-lower-bound}
Let $\epsilon\in[1/n,1]$, $\delta\leq \frac{1}{16}(1-e^{-\epsilon})$. 
For any $(\epsilon,\delta)$-DP algorithm, $\cA$, there exists a dataset $S$ with empirical minimizer of norm $B$ 
such that the excess empirical risk of $\cA$ is lower bounded by $\Omega\bc{\min\bc{\ybound^2,\frac{(B\xbound)^{4/3}(H\ybound)^{2/3}}{(n\epsilon)^{2/3}}, \frac{\sqrt{\dimension}B\xbound\ybound\sqrt{H}}{n\epsilon}}}$.
\end{theorem}
The problem instance used in the proof is the squared loss function, and thus this result holds additionally for the more specific case of linear regression.
We also note this lower bound implies our upper bound is optimal whenever $B\xbound \leq \ybound$, which is a commonly studied regime in linear regression \cite{SST10,koren2015fast}.
We here provide a proof sketch and defer the full proof to Appendix \ref{app:smooth-lowerbound}.
\paragraph{Proof sketch of Theorem \ref{thm:smooth-lower-bound}}
Define $F(w;S) = \frac{1}{n}\sum_{(x,y)\in S} (\ip{w}{x}-y)^2$.
Let $d'<\min\bc{n,\dimension}$
and $b,p \in [0,1]$ 
be parameters to be chosen later. For any  $\bssigma \in \bc{\pm 1}^{d'}$, define the dataset $S_{\bssigma}$ which consists of the union of $d'$ subdatasets, $S_1,...,S_{d'}$ given as follows. Set $\frac{pn}{d'}$ of the feature vectors in $S_{j}$ 
as $\scale e_j$ (the rescaled $j$'th standard basis vector) and the rest as the zero vector. Set $\frac{pn}{2d}(1+b)$ of the labels as $\bssigma_j \ybound$ and $\frac{pn}{2d}(1-b)$ labels as $-\bssigma_j \ybound$. 
Let $w^{\bssigma} = \argmin_{w\in\re^d}\bc{F(w;S_{\bssigma})}$ be the ERM minimizer of $F(\cdot;S_{\bssigma})$.
Following from Lemma 2 of \cite{shamir_2015} we have that for any $\bar{w} \in \re^{d}$ that
\begin{equation} %
F(\bar{w};S_{\bssigma}) - F(w^{\bssigma};S_{\bssigma}) \geq \frac{p\scale^2}{2d'}\sum_{j=1}^{d'}{(\bar{w_j} - w^{\bssigma}_j)^2}. 
\end{equation} 
We will now show lower bounds on the per-coordinate error. 
Consider any $\bssigma$ and $\bssigma'$ which
differ only at index $j$ for some $j\in[d']$.
Note that the datasets $S_{\bssigma}$ and $S_{\bssigma'}$  differ in $\Delta=\frac{pn}{2d'}[(1+b)-(1-b)] = \frac{pbn}{d'}$ points.
Let $\tau = w_j^{\bssigma} = \frac{\ybound b}{\xbound}$ and $\tau' = w_j^{\bssigma'} = -\frac{\ybound b}{\xbound}$ (i.e. the $j$ components of the empirical minimizers for $S$ and $S'_j$ respectively).
Note that $|w_j^{\bssigma} - w_j^{\bssigma'}| = \frac{2\ybound b}{\xbound}$. 
Setting $d' = \big(\frac{p\scale Bn\epsilon}{\ybound}\big)^{2/3}$ and
$b = \big(\frac{\scale B}{\ybound \sqrt{pn \epsilon}}\big)^{2/3}$ ensures $\Delta \leq \frac{1}{\epsilon}$, and thus
Lemma \ref{lem:ch} can be used to obtain
\begin{align*} \label{eq:ybaH_bound-main}
\ex{}{|\cA(S_{\bssigma})_j - w_j^{\bssigma}|^2 + |\cA(S_{\bssigma'})_j - w_j^{\bssigma'}|^2} \geq
\frac{1}{32}\frac{\ybound^2 b^2}{\xbound^2} =
\frac{B^{4/3}\ybound^{2/3}}{32(\scale pn\epsilon)^{2/3}}.
\end{align*}

One can now show via a packing argument that 
\begin{align*}
\sup_{\bssigma \in \bc{\pm 1}^{d'}}\bc{\ex{}{F(\cA(S_{\bssigma})) - F(w^{\bssigma})}} 
&\geq \frac{(\scale B)^{4/3}\ybound^{2/3}p^{1/3}}{128 (n\epsilon)^{2/3}},
\end{align*}
The result then follows from setting $p= \min\bc{1,\frac{d^{3/2}\ybound}{B\scale n\epsilon}}$.

\section{Lipschitz GLMs} \label{sec:lip_glm}
In the Lipschitz case, we close a subtle gap in existing rates. We recall that in this setting a more precise characterization in terms of the expected rank of the design matrix is possible (as opposed to using $d$). The best known upper bound is $\tilde{O}\br{\frac{\mnorm \sqrt{\rank}}{n\epsilon}}$ assuming knowledge of $\mnorm$. This bound was shown to be optimal when $\epsilon \geq \rank/n$ and $\mnorm=\sqrt{\rank}$ \cite{SSTT21}. 

We first show that in the high privacy regime where $\epsilon \leq \rank/n$, an improved rate is possible. 
Specifically, we show that in this regime constrained regularized ERM with output perturbation and achieves the optimal rate. 
In fact, we note that the method of \cite{JainThakurta:2014} (i.e.~\emph{unconstrained} regularized ERM with output perturbation), can obtain this rate when $\epsilon = O(1)$ if the regularization parameter is set differently. We present the constrained version in order to leverage Rademacher complexity arguments and provide a slightly cleaner bound that holds for all $\epsilon>0$. 
\begin{theorem} \label{thm:lipschitz-output-perturbation}
Algorithm \ref{alg:constrained-reg-erm-output-perturbation} run with parameters $\sigma^2 = \frac{4G^2\norm{\cX}^2\log{1/\delta}}{\lambda^2n^2\epsilon^2}$ and $\lambda = \frac{G\xbound\br{\log{1/\delta}}^{1/4}}{B\sqrt{n\epsilon}}$ satisfies $(\epsilon,\delta)$-DP. Given a dataset of $n$ i.i.d.~samples from $\cD$, its output has excess risk
$    \ex{}{\excessrisk(\out)} =\tilde{O}\br{\frac{GB\xbound}{\sqrt{n}} + \frac{GB\xbound}{\sqrt{n\epsilon}}}$.
\end{theorem}

We also state and prove a similar bound using the JL technique in Appendix \ref{app:lip_glm}.

Next, we generalize the lower bound of \cite{SSTT21} to show this new bound is optimal for all settings of $B$, $\rank$, and $\epsilon$. 
We now show that a modification of the lower bound present in \cite{SSTT21} shows our upper bound is tight. We note that their lower bound only held for problem instances where $\rank=\mnorm^2$ and $\epsilon \leq \rank/n$. By contrast, the upper bound $O(\frac{\sqrt{\rank} \mnorm}{n\epsilon})$ holds for any values of $\rank$ and $\mnorm$. 
\begin{theorem} \label{thm:lipschitz_lower}
Let $G,\norm{\cY},\norm{\cX}, B>0$, $\epsilon \leq 1.2$ and $\delta \leq \epsilon$. For any $(\epsilon,\delta)$-DP algorithm $\cA$, there exists a $G$-Lipschitz GLM loss bounded at zero by $\norm{\cY}$ and a distribution $\cD$ with $\mnorm \leq B$ such that the output of $\cA$ on $S\sim\cD^n$ 
satisfies
$    \mathbb{E}[\excessrisk(\cA(S))] =\Omega\Big(GB\xbound\min\Big(1,\frac{1}{\sqrt{n\epsilon}},\frac{\sqrt{\rank}}{n\epsilon}\Big)\Big)$.
\end{theorem}
All proofs for this section are deferred to Appendix \ref{app:lip_glm}.
\section{Adapting to \texorpdfstring{$\mnorm$}{}} \label{sec:adapt}
Our method for privately adapting to $\mnorm$ is given in Algorithm \ref{alg:adapt}. We start by giving a high level overview and defining some necessary preliminaries. The algorithm works in the following manner. First we define a number of ``guesses'' $K$ for $\mnorm$, $B_1,...,B_K$ where $B_j=2^j:\forall j\in[K]$. Then given black box access to a DP optimization algorithm, $\cA$, Algorithm \ref{alg:adapt} generates $K$ candidate vectors $w_1,...,w_K$ using $\cA$, training set $S_1\in(\cX\times\cY)^{n/2}$, and the guesses $B_1,...,B_K$. We assume $\cA$ satisfies the following accuracy assumption for some confidence parameter $\beta>0$.
\begin{assumption} \label{asm:risk_bound}
There exists a function $\err:\re^+\mapsto\re^+$ such that for any $B\in\re^+$,
whenever $B\geq\mnorm$, w.p. at least $1-\frac{\beta}{4K}$ under the randomness of $S_1\sim\cD^{\frac{n}{2}}$ and $\cA$ it holds that
$\excessrisk(\cA(S_1,B);\cD) \leq \err(B)$.
\end{assumption}
After generating the candidate vectors, the goal is to pick guess with the smallest excess population risk in a differentially private manner using a validation set $S_2$. 
The following assumption on $\cA$ allows us both to ensure the privacy of the model selection algorithm and verify that $\eloss(w_j;S_2)$ provides a tight estimate of $L(w_j;\cD)$.
\begin{assumption} \label{asm:loss_bound}
There exist a function $\Delta:\re^+\mapsto\re^+$ such that for any dataset $S_2 \in (\cX\times\cY)^{n/2}$ and $B>0$
\begin{equation}
\underset{\cA}{\mathbb{P}}[\exists (x,y)\in S_2: |\ell(\cA(S_1,B);(x,y))|
    \geq \Delta(B)] \leq \frac{\min\bc{\delta,\beta}}{4K}\nonumber
\end{equation}
\end{assumption}
Specifically, our strategy will be to use the Generalized Exponential Mechanism, $\gem$, of \cite{RS15} in conjunction with a penalized score function. 
Roughly, this score function penalty ensures the looser guarantees on the population loss estimate when $B$ is large do not interfere with the loss estimates at smaller values of $B$. We provide the relevant details for $\gem$ in Appendix \ref{app:gem}. 
We now state our result.

\begin{algorithm}[t]
\caption{Private Grid Search}
\label{alg:adapt}
\begin{algorithmic}[1]
\REQUIRE Dataset $\cS \in (\cX \times \cY)^n$, grid parameter $K \in \re$, optimization algorithm: $\cA:(\cX \times \cY)^n \times \re \mapsto \re^d$, 
privacy parameters $(\epsilon, \delta)$

\STATE Partition $S$ into two disjoint sets, 
$S_1$ and $S_2$, of size $\frac{n}{2}$%

\STATE $w_0 = 0$

\FOR{$j \in [K]$}

\STATE $B_j = 2^j$

\STATE $w_j = \cA(S_1, B_j)$ 

\STATE $\tilde{L}_j = \eloss(w_j;S_2) + \frac{\Delta(B_j)\log{K/\beta}}{n} + \sqrt{\frac{4\ybound^2\log{K/\beta}}{n}}$ 

\ENDFOR

\STATE Set $j^*$ as the output of $\gem$ run with privacy parameter $\frac{\epsilon}{2}$, confidence parameter $\frac{\beta}{4}$, and sensitivity/score pairs $(0,\ybound^2),(\Delta(B_1),\tilde{L}_1)...,(\Delta(B_K),\tilde{L}_K)$,

\STATE Output $w_{j^*}$
\end{algorithmic}
\end{algorithm}

\begin{theorem} \label{thm:adapt}
Let $\ell:\re^d\times(\cX\times\cY)$ be a smooth non-negative loss function such that $\ell(0,(x,y))\leq \ybound^2$ for any $x,y\in(\cX\times\cY)$. Let $\epsilon,\delta,\beta\in[0,1]$.
Let $K>0$ satisfy $\err(2^{K})\geq\ybound^2$.
Let $\cA$ be an $(\frac{\epsilon}{2K},\frac{\delta}{2K})$-DP algorithm satisfying Assumption \ref{asm:loss_bound}. 
Then Algorithm \ref{alg:adapt} is $(\epsilon,\delta)$-DP. Further, if $\cA$ satisfies Assumption \ref{asm:risk_bound} and $S_1\sim \cD^{n/2}$ 
then Algorithm \ref{alg:adapt} outputs $\bar{w}$ s.t. with probability at least $1-\beta$,
\begin{align*}
\excessrisk(&\bar{w};\cD) \leq \min\Big\{\ybound^2, 
\err(2\max\bc{\norm{w^*},1}) 
+ \sqrt{\frac{4\ybound^2\log{4K/\beta}}{n}}
+ \frac{5\Delta(2\max\bc{\mnorm,1})}{n\epsilon}\Big\}.
\end{align*}
\end{theorem} 

We note that we develop a generic confidence boosting approach to obtain high probability guarantees from our previously described algorithms in Section \ref{sec:boost}, and thus obtaining algorithms which satisfy \ref{asm:risk_bound} is straightforward. We provide more details on how our algorithms satisfy Assumption \ref{asm:loss_bound} in Appendix \ref{app:stab-asm}.
The following Theorem details the guarantees implied by this method for output perturbation with boosting (see Theorems \ref{thm:boosting},\ref{thm:boosted-smooth-output-perturbation}). Full details are in Appendix \ref{app:pf-smooth-reg-pert-adapt}.
\begin{theorem} \label{thm:smooth-reg-pert-adapt}
Let $K,\epsilon,\delta,\beta>0$ and $\cA$ be the algorithm formed by running \ref{alg:constrained-reg-erm-output-perturbation} with boosting and privacy parameters $\epsilon'=\frac{\epsilon}{K}$, $\delta'=\frac{\delta}{K}$. 
Then there exists a setting of $K$ such that $K=\Theta\br{\log{\max\bc{\frac{\ybound\sqrt{n}}{\xbound\sqrt{H}},\frac{\ybound^2(n\epsilon)^{2/3}}{\sqrt{H}\xbound^2}}}}$ and Algorithm \ref{alg:adapt} run with $\cA$ and $K$ is $(\epsilon,\delta)$-DP and when given $S\sim\cD^n$, satisfies the following w.p. at least $1-\beta$ (letting $B^* = 2\max\bc{\mnorm,1}$)
\begin{align*}
    \excessrisk(\out) &= \tilde{O}\Bigg(\min\Big\{\ybound^2,
    \frac{\br{\sqrt{H}B^*\norm{\cX}}^{4/3}\norm{\cY}^{2/3}+ \br{\sqrt{H}B^*\norm{\cX}}^2}{(n\epsilon)^{2/3}} \\
    &+\frac{\sqrt{H}B^*\norm{\cX}\max\bc{\norm{\cY},1} + \norm{\cY}^2}{\sqrt{n}} + \frac{\ybound^2 + H(B^*\xbound)^2}{n\epsilon}\Big\}\Bigg).
\end{align*}
\end{theorem}

\vspace{-10pt}
\label{sec:boost}
\paragraph{Confidence Boosting:}
We give an algorithm to boost the confidence of unconstrained, smooth DP-SCO (with possibly non-Lipschitz losses).
We split the dataset $S$ into $m+1$ chunks and run an $(\epsilon,\delta)$-DP algorithm over the $m$ chunks to get $m$ models, and then use Report Noisy Max mechanism to select a model with approximately the least empirical risk.
We show that this achieves the optimal rate of $\tilde O\br{\frac{\sqrt{H}B\xbound\ybound}{\sqrt{n}}}$ whereas the previous high probability result of 
\cite{SST10} had an additional
$\tilde O\br{\frac{\ybound^2}{\sqrt{n}}}$ term, which was also limited to only GLMs.
The key idea is that non-negativity, convexity, smoothness and loss bounded at zero, all together enable strong bounds on the variance of the loss, and consequently give stronger concentration bounds. 
The details are deferred to Appendix \ref{sec:appendix_boosting}.

\section*{Acknowledgements}\label{sec:ack}
RA and EU are supported, in part, by NSF BIGDATA award IIS-1838139 and NSF CAREER award IIS-1943251. RB's and MM's work on this research is supported by NSF Award AF-1908281 and Google Faculty Research Award. CG's research is partially supported by
INRIA Associate Teams project, FONDECYT 1210362 grant, and National Center for Artificial Intelligence CENIA FB210017, Basal ANID. Part of this work was done while CG was at the University of Twente.

\appendix

\bibliography{arxiv/main}

\newcommand{\etalchar}[1]{$^{#1}$}
\begin{thebibliography}{BFTGT19}

\bibitem[ACKL21]{Amir:2021never}
Idan Amir, Yair Carmon, Tomer Koren, and Roi Livni.
\newblock Never go full batch (in stochastic convex optimization).
\newblock {\em Advances in Neural Information Processing Systems}, 34, 2021.

\bibitem[AKL21]{Amir:2021sgd}
Idan Amir, Tomer Koren, and Roi Livni.
\newblock Sgd generalizes better than gd (and regularization doesn’t help).
\newblock In {\em Proceedings of Thirty Fourth Conference on Learning Theory},
  volume 134 of {\em Proceedings of Machine Learning Research}, pages 63--92.
  PMLR, 15--19 Aug 2021.

\bibitem[BBL03]{bousquet2003introduction}
Olivier Bousquet, St{\'e}phane Boucheron, and G{\'a}bor Lugosi.
\newblock Introduction to statistical learning theory.
\newblock In {\em Summer school on machine learning}, pages 169--207. Springer,
  2003.

\bibitem[BE02]{Bousquet:2002}
Olivier Bousquet and Andr\'{e} Elisseeff.
\newblock Stability and generalization.
\newblock {\em J. Mach. Learn. Res.}, 2:499–526, March 2002.

\bibitem[BFGT20]{bassily2020stability}
Raef Bassily, Vitaly Feldman, Crist{\'o}bal Guzm{\'a}n, and Kunal Talwar.
\newblock Stability of stochastic gradient descent on nonsmooth convex losses.
\newblock {\em Advances in Neural Information Processing Systems}, 33, 2020.

\bibitem[BFTGT19]{bassily2019private}
Raef Bassily, Vitaly Feldman, Kunal Talwar, and Abhradeep Guha~Thakurta.
\newblock Private stochastic convex optimization with optimal rates.
\newblock In {\em Advances in Neural Information Processing Systems},
  volume~32. Curran Associates, Inc., 2019.

\bibitem[BGN21]{bassily2021non}
Raef Bassily, Crist{\'o}bal Guzm{\'a}n, and Anupama Nandi.
\newblock Non-euclidean differentially private stochastic convex optimization.
\newblock In {\em Conference on Learning Theory}, pages 474--499. PMLR, 2021.

\bibitem[BLM13]{boucheron2013concentration}
St{\'e}phane Boucheron, G{\'a}bor Lugosi, and Pascal Massart.
\newblock {\em Concentration inequalities: A nonasymptotic theory of
  independence}.
\newblock Oxford university press, 2013.

\bibitem[BST14]{bassily2014private}
Raef Bassily, Adam Smith, and Abhradeep Thakurta.
\newblock Private empirical risk minimization: Efficient algorithms and tight
  error bounds.
\newblock In {\em 2014 IEEE 55th Annual Symposium on Foundations of Computer
  Science}, pages 464--473. IEEE, 2014.

\bibitem[CH12]{CH12}
Kamalika Chaudhuri and Daniel~J. Hsu.
\newblock Convergence rates for differentially private statistical estimation.
\newblock In {\em Proceedings of the 29th International Conference on Machine
  Learning, {ICML} 2012, Edinburgh, Scotland, UK, June 26 - July 1, 2012}.
  icml.cc / Omnipress, 2012.

\bibitem[CWH20]{Xiangyi_clipping}
Xiangyi Chen, Steven~Z. Wu, and Mingyi Hong.
\newblock Understanding gradient clipping in private sgd: A geometric
  perspective.
\newblock In H.~Larochelle, M.~Ranzato, R.~Hadsell, M.~F. Balcan, and H.~Lin,
  editors, {\em Advances in Neural Information Processing Systems}, volume~33,
  pages 13773--13782. Curran Associates, Inc., 2020.

\bibitem[CWZ21]{CWZ20}
T.~Tony Cai, Yichen Wang, and Linjun Zhang.
\newblock {The cost of privacy: Optimal rates of convergence for parameter
  estimation with differential privacy}.
\newblock {\em The Annals of Statistics}, 49(5):2825 -- 2850, 2021.

\bibitem[DKM{\etalchar{+}}06]{DKMMN06}
Cynthia Dwork, Krishnaram Kenthapadi, Frank McSherry, Ilya Mironov, and Moni
  Naor.
\newblock Our data, ourselves: Privacy via distributed noise generation.
\newblock In {\em EUROCRYPT}, 2006.

\bibitem[DR{\etalchar{+}}14]{dwork2014algorithmic}
Cynthia Dwork, Aaron Roth, et~al.
\newblock The algorithmic foundations of differential privacy.
\newblock {\em Found. Trends Theor. Comput. Sci.}, 9(3-4):211--407, 2014.

\bibitem[FKT20]{FKT:2020}
Vitaly Feldman, Tomer Koren, and Kunal Talwar.
\newblock Private stochastic convex optimization: optimal rates in linear time.
\newblock In {\em Proceedings of the 52nd Annual ACM SIGACT Symposium on Theory
  of Computing}, pages 439--449, 2020.

\bibitem[HRS15]{HRS15}
Moritz Hardt, Benjamin Recht, and Yoram Singer.
\newblock Train faster, generalize better: Stability of stochastic gradient
  descent.
\newblock {\em CoRR}, abs/1509.01240, 2015.

\bibitem[JT14]{JainThakurta:2014}
Prateek Jain and Abhradeep~Guha Thakurta.
\newblock (near) dimension independent risk bounds for differentially private
  learning.
\newblock In Eric~P. Xing and Tony Jebara, editors, {\em Proceedings of the
  31st International Conference on Machine Learning}, volume~32 of {\em
  Proceedings of Machine Learning Research}, pages 476--484, Bejing, China,
  22--24 Jun 2014. PMLR.

\bibitem[KL15]{koren2015fast}
Tomer Koren and Kfir~Y Levy.
\newblock Fast rates for exp-concave empirical risk minimization.
\newblock In {\em NIPS}, pages 1477--1485, 2015.

\bibitem[Nel11]{Nelson2011SketchingAS}
Jelani Nelson.
\newblock {\em Sketching and streaming high-dimensional vectors}.
\newblock PhD thesis, Massachusetts Institute of Technology, 2011.

\bibitem[N{\^{e}}UZ20]{NUZ20}
Huy~L\^{e} Nguy\~{\^{e}}n, Jonathan Ullman, and Lydia Zakynthinou.
\newblock Efficient {P}rivate {A}lgorithms for {L}earning {L}arge-{M}argin
  {H}alfspaces.
\newblock In Aryeh Kontorovich and Gergely Neu, editors, {\em Proceedings of
  the 31st International Conference on Algorithmic Learning Theory}, volume 117
  of {\em Proceedings of Machine Learning Research}, pages 704--724. PMLR, 08
  Feb--11 Feb 2020.

\bibitem[NY83]{Nemirovsky:1983}
A.S. Nemirovsky and E.R. Yudin.
\newblock {\em Problem Complexity and Method Efficiency in Optimization}.
\newblock A Wiley-Interscience publication. Wiley, 1983.

\bibitem[RS15]{RS15}
Sofya Raskhodnikova and Adam~D. Smith.
\newblock Efficient lipschitz extensions for high-dimensional graph statistics
  and node private degree distributions.
\newblock {\em CoRR}, abs/1504.07912, 2015.

\bibitem[Sha15]{shamir_2015}
Ohad Shamir.
\newblock The sample complexity of learning linear predictors with the squared
  loss.
\newblock {\em J. Mach. Learn. Res.}, 16:3475--3486, 2015.

\bibitem[SSBD14]{shalev2014understanding}
Shai Shalev-Shwartz and Shai Ben-David.
\newblock {\em Understanding machine learning: From theory to algorithms}.
\newblock Cambridge university press, 2014.

\bibitem[SST10]{SST10}
Nathan Srebro, Karthik Sridharan, and Ambuj Tewari.
\newblock Smoothness, low noise and fast rates.
\newblock In J.~Lafferty, C.~Williams, J.~Shawe-Taylor, R.~Zemel, and
  A.~Culotta, editors, {\em Advances in Neural Information Processing Systems},
  volume~23. Curran Associates, Inc., 2010.

\bibitem[SSTT20]{SSTT21}
Shuang Song, Thomas Steinke, Om~Thakkar, and Abhradeep Thakurta.
\newblock Characterizing private clipped gradient descent on convex generalized
  linear problems.
\newblock {\em CoRR}, abs/2006.06783, 2020.

\bibitem[SU17]{steinke2017tight}
Thomas Steinke and Jonathan Ullman.
\newblock Tight lower bounds for differentially private selection.
\newblock In {\em 2017 IEEE 58th Annual Symposium on Foundations of Computer
  Science (FOCS)}, pages 552--563. IEEE, 2017.

\bibitem[Ver18]{vershynin2018high}
Roman Vershynin.
\newblock {\em High-dimensional probability: An introduction with applications
  in data science}, volume~47.
\newblock Cambridge university press, 2018.

\bibitem[VGNA20]{vladimirova2020sub}
Mariia Vladimirova, St{\'e}phane Girard, Hien Nguyen, and Julyan Arbel.
\newblock Sub-weibull distributions: Generalizing sub-gaussian and
  sub-exponential properties to heavier tailed distributions.
\newblock {\em Stat}, 9(1):e318, 2020.

\bibitem[Wan18]{WGS18}
Yu{-}Xiang Wang.
\newblock Revisiting differentially private linear regression: optimal and
  adaptive prediction {\&} estimation in unbounded domain.
\newblock In Amir Globerson and Ricardo Silva, editors, {\em Proceedings of the
  Thirty-Fourth Conference on Uncertainty in Artificial Intelligence, {UAI}
  2018, Monterey, California, USA, August 6-10, 2018}, pages 93--103. {AUAI}
  Press, 2018.

\end{thebibliography}
\bibliographystyle{alpha}

\newpage
\appendix

\section{Missing Proofs from Section \ref{sec:smooth_upper} (Smooth GLMs) }
\subsection{Utility Lemmas}
\begin{fact}\cite{shalev2014understanding}
\label{fact:self-boundedness}
For a $\tilde H$-smooth non-negative function $f$, and for any $u \in \operatorname{dom}(f)$, we have $\norm{\nabla f(u)} \leq  \sqrt{4\tilde H f(u)}$. 
\end{fact}

\subsection{Proof of Lemma \ref{lem:smooth_lip}}
From the self-bounding property (Fact \ref{fact:self-boundedness}), the $\ybound^2$ bound on loss at zero, and smoothness, we have the following bound on the gradient:
\begin{align}
\label{eqn:gradient-bound-self-bounding}
\nonumber
    \norm{\nabla \ell(w;(x,y)} &\leq \norm{\nabla \ell(0;(x,y))} + \norm{\nabla \ell(w;(x,y)) - \nabla \ell(0;(x,y)) } \\
    \nonumber
    &\leq  2\sqrt{H}\norm{x} \ell(0;(x,y)) + H\norm{x}^2\norm{w}\\
    &\leq 2\br{\sqrt{H}\norm{\cY}\norm{x} + H\norm{x}^2\norm{w}}.
\end{align}

\begin{lemma} \label{lem:smooth_loss_bound}
For any $w\in\ball{B}$ and any $(x,y)\in(\cX\times\cY)$ it holds that $\ell(w;(x,y)) \leq 3(\ybound^2 + HB^2\xbound^2)$.
\end{lemma}
\begin{proof}
Using the fact that the loss function is $G = \ybound\sqrt{H}\xbound + BH\xbound^2$-Lipschitz in the constraint set (Lemma \ref{lem:smooth_lip}) we have
\begin{align*}
    \ell(w;(x,y)) &\leq \ell(0;(x,y)) + \abs{\ell(w;(x,y))-\ell(0;(x,y))} \\
    &\leq \ybound^2 + G\norm{w} \\
    &\leq \ybound^2 + \ybound\sqrt{H}B\xbound + HB^2\xbound^2\\ &\leq  3\br{\ybound^2 +HB^2\xbound^2}
\end{align*}
where the last step follow from AM-GM inequality.
\end{proof}
\subsection{Low Dimension}
Before presenting the proof of Theorem \ref{thm:smooth-low-dim}, we provide formal statements of its Corollaries.

Corollary \ref{cor:smooth-low-dim-non-priv}, stated below, gives an upper bound on excess risk of gradient descent in the non-private setting.

\begin{corollary}
\label{cor:smooth-low-dim-non-priv}
Let $\ell$ be a non-negative convex $\tilde H$ smooth loss function, bounded at zero by $\norm{\cY}^2$. 
Let $n_0 = \frac{\tilde HB^2}{\ybound^2}$,
$\cW=\ball{B}$, $T=n$, and $\eta = \min\Big(\frac{B}{\sqrt{T}\sqrt{\tilde H}\ybound}, \frac{1}{4\tilde H}\Big)$. 
Given a dataset $S$ of $n\geq n_0$ i.i.d samples from an unknown distribution $\cD$, the excess risk of output of Algorithm \ref{alg:noisySGD} with $\sigma=0$ is bounded as,
\begin{align*}
    \mathbb{E}[L(\hat w;\cD) - L(w^*;\cD)] \leq O\br{
    \frac{\sqrt{\tilde H}B\ybound}{\sqrt{n}}}.
\end{align*}
\end{corollary}

Corollary \ref{cor:smooth_low_dim_glm} below gives an upper bound on excess risk of noisy gradient descent for non-negative smooth GLMs.

\begin{corollary} \label{cor:smooth_low_dim_glm}
Let $n_0 = \frac{H\xbound^2B^2}{\ybound^2}$,
$\eta = \min\Big(\frac{B}{\sqrt{T}\max\br{\sqrt{ H}\xbound\ybound, \sigma\sqrt{d}}}, \frac{1}{4H \xbound}\Big)$,  $\cW = \ball{B}$, $\sigma^2 = \frac{8G^2T\log{1/\delta}}{n^2\epsilon^2}$  and 
$T=n$.
Let $\ell$ be a non-negative convex $H$ smooth GLM, bounded at zero by $\norm{\cY}^2$. 
Algorithm \ref{alg:noisySGD} satisfies $(\epsilon, \delta)$-differential privacy.
Given a dataset $S\sim\cD^n$, $n\geq n_0$, then
the excess risk of output of Algorithm \ref{alg:noisySGD} is bounded as,
\begin{align*}
    \mathbb{E}[L(\hat w;\cD) - L(w^*;\cD)] 
    = O\Bigg(
    \frac{\sqrt{H}\xbound B\ybound}{\sqrt{n}} 
    + \frac{\br{\sqrt{H}\xbound\ybound + H \xbound^2B^2}B\sqrt{d \log{1/\delta}}}{n\epsilon}\Bigg).
\end{align*}
\end{corollary}

\subsubsection{Proof of Theorem \ref{thm:smooth-low-dim}}
\label{sec:app_smooth_upperbounds_lowd}
Since Algorithm \ref{alg:noisySGD} uses a projection step, the iterates always lie in the constraint set $\bc{w:\norm{w}\leq B}$. Hence, the function over this constraint set is $G$-Lipschitz. From the analysis of Noisy (S)GD in \cite{bassily2014private,bassily2019private}, we have that the setting of noise variance $\sigma^2$ ensures that the algorithm satisfies $(\epsilon,\delta)$-DP

We now move to the utility part. 
We start with the decomposition of excess risk as
\begin{align}
\label{eqn:jl-smooth-noisy-gd-excess-risk}
    \mathbb{E}[L(\hat w;\cD) - L(w^*; \cD)] =  \mathbb{E}[L(\hat w;\cD) - \hat L(\hat w; S)] +  \mathbb{E}[\hat L(\hat w;S) - \hat L( w^*; S)] 
\end{align}

The key arguments are as follows: we first bound generalization gap, or on-average stability (first term in the right hand side above), in terms of average argument stability and excess empirical risk (Lemma \ref{lem:gen-gap-uas}). We then bound average argument stability in terms of average regret (Lemma \ref{lem:gd-average-stability}).
Finally, in Lemma \ref{lem:noisy-sgd-erm}, we provide bounds on excess empirical risk and average regret of gradient descent. Substituting these in the above equation gives the claimed bound.
We now fill in the details.

We start with Lemma \ref{lem:gen-gap-uas} which gives the following bound on the generalization gap:
\begin{align}
\label{eq:smooth-gen-gap-eqn}
    \mathbb{E}[L(\cA(S);\cD) - \hat L(\hat w; S)] &\leq   \frac{\sqrt{\tilde H}B}{\sqrt{n}\norm{\cY}}\br{\mathbb{E}[\hat L(\hat w;S) - \hat L(w^*;S)]} + 
    \frac{2\sqrt{\tilde H n}\ybound}{B}\uas^2
    + \frac{\sqrt{\tilde H}B\ybound}{\sqrt{n}}
\end{align}   

Substituting the bound on $\uas$ from Lemma \ref{lem:gd-average-stability}, the second term becomes,
\begin{align*}
     \frac{2\sqrt{\tilde H n}\ybound}{B}\uas^2 &\leq 
     \frac{ 16\tilde H^{3/2}\eta^2T\ybound}{\sqrt{n}B}
     \frac{1}{n}\sum_{j=1}^T \mathbb{E}[\eloss(w_j;S) - \eloss(w^*;S)] +      \frac{ 16\tilde H^{3/2}\eta^2T\ybound}{\sqrt{n}B}\\
     &\leq \frac{16\sqrt{\tilde H}B}{\sqrt{n}\ybound} \frac{1}{n}\sum_{j=1}^T \mathbb{E}[\eloss(w_j;S) - \eloss(w^*;S)] +  \frac{16\sqrt{\tilde H}B}{\sqrt{n}\ybound}\\
     & \leq \frac{16}{n}\sum_{j=1}^T \mathbb{E}[\eloss(w_j;S) - \eloss(w^*;S)] +  \frac{16\sqrt{\tilde H}B}{\sqrt{n}\ybound}.
\end{align*}
Substituting the above in Eqn. \eqref{eq:smooth-gen-gap-eqn} and using the fact that $\frac{\sqrt{\tilde H}B}{\sqrt{n}\norm{\cY}} \leq 1$ from the lower bound on $n$, we get:
\begin{align*}
    \mathbb{E}[L(\cA(S);\cD) - \hat L(\hat w; S)] &\leq 
      \br{\mathbb{E}[\hat L(\hat w;S) - \hat L(w^*;S)]} + 
 \frac{16}{n}\br{\sum_{j=1}^T \mathbb{E}[\eloss(w_j;S) - \eloss(w^*;S)]}
   \\& +  \frac{17\sqrt{\tilde H}B\ybound}{\sqrt{n}}.
\end{align*}

From the excess empirical risk guarantee (Lemma \ref{lem:noisy-sgd-erm}), the terms
excess empirical risk $\mathbb{E}[\hat L(\hat w;S) - \hat L(w^*;S)]$ and average regret $\frac{1}{n}\br{\sum_{j=1}^T \mathbb{E}[\eloss(w_j;S) - \eloss(w^*;S)]}$ are both bounded by the same quantity. 
Thus, substituting the above in Eqn. \eqref{eqn:jl-smooth-noisy-gd-excess-risk} and substituting the bound from \ref{lem:noisy-sgd-erm}, we have, 
\begin{align*}
      \mathbb{E}[L(\hat w;\cD) - L(w^*; \cD)]  &\leq \frac{18}{n}\sum_{j=1}^n\mathbb{E}[\hat L(w_j;S)-\hat L(w^*;S)] + \frac{17\sqrt{\tilde H}B\ybound}{\sqrt{n}} \\
      &\leq O\br{\frac{\sqrt{H}B\ybound}{\sqrt{n}} + \frac{\sqrt{d}GB\log{1/\delta}}{n\epsilon}}.
\end{align*}

Substituting the value of $G$ completes the proof.

\begin{lemma}
\label{lem:gen-gap-uas}
Let $n\geq \frac{\tilde H B^2}{\ybound^2}$.
Let $\ell$ be a non-negative $\tilde H$ smooth convex loss function.
Let $S$ be a dataset of $n$ i.i.d. samples from an unknown distribution $\cD$, and $w^*$ denote the optimal population risk minimizer.
The generalization gap of algorithm $\cA$ is bounded as,

\begin{align*}
    \mathbb{E}[L(\cA(S);\cD) - \hat L(\cA(S); S)] &\leq   \frac{\sqrt{\tilde H}B}{\sqrt{n}\norm{\cY}}\br{\mathbb{E}[\hat L(\cA(S);S) - \hat L(w^*;S)]} + 
    \frac{2\sqrt{\tilde H n}\ybound}{B}\uas^2
    \\&+ \frac{\sqrt{\tilde H}B\ybound}{\sqrt{n}}.
\end{align*}    

\end{lemma}

\begin{proof}
Let $\hat w:=\cA(S)$,
 $S^{(i)}$ be the dataset where the $i$-th data point is replaced by an i.i.d. point $(x',y')$ and let $\hat w^{(i)}$ be the corresponding output of $\cA$.

A standard fact (see \cite{shalev2014understanding}) is that generalization gap is equal to on-average stability:
\begin{align*}
    \mathbb{E}[L(\hat w;\cD) - \hat L(\hat w; S)] =\mathbb{E}[\ell(  \hat w^{(i)}; (x_i,y_i)) - \ell(\hat w; (x_i,y_i)].
\end{align*}

From smoothness and self-bounding property, we have,
\begin{align*}
    \ell(  \hat w^{(i)}; (x_i,y_i)) - \ell(\hat w; (x_i,y_i) &\leq \norm{\nabla \ell(\hat w;(x_i,y_i))}\norm{\hat w - \hat w^{(i)}} + \frac{\tilde H}{2}\norm{\hat w - \hat w^{(i)}}^2 \\
    &\leq 2\sqrt{\tilde H\ell((\hat w;(x_i,y_i))}\norm{\hat w - \hat w^{(i)}} + \frac{\tilde H}{2}\norm{\hat w - \hat w^{(i)}}^2.
\end{align*}
Taking expectation, using Cauchy-Schwarz inequality and substituting average argument stability, we get, 
\begin{align}
\label{eqn:gen-gap-average-stability-eqn}
\nonumber
    &\mathbb{E}\left[\ell(  \hat w^{(i)}; (x_i,y_i)) - \ell(\hat w; (x_i,y_i)\right] \\
    &\leq 2 \sqrt{\tilde H\mathbb{E}[\ell((\hat w;(x_i,y_i))]}\sqrt{\mathbb{E}[\norm{\hat w - \hat w^{(i)}}^2]} + \frac{\tilde H}{2}\mathbb{E}\left[\norm{\hat w - \hat w^{(i)}}^2\right] \\
    \nonumber
    &\leq 2\sqrt{\tilde H\mathbb{E}[\hat L(\hat w;S)]}
    \uas + \frac{\tilde H\uas^2}{2}\\
    \nonumber
    &\leq \frac{\sqrt{\tilde H}B\mathbb{E}[\hat L(\hat w;S)]}{\sqrt{n}\norm{\cY}} 
    + \frac{\sqrt{\tilde H n }\norm{\cY}\uas^2}{B} + \frac{\tilde H\uas^2}{2} 
    \\
    \nonumber
    &\leq \frac{\sqrt{\tilde H}B}{\sqrt{n}\norm{\cY}}\br{\mathbb{E}[\hat L(\hat w;S) - \hat L(w^*;S)]} + \frac{\sqrt{\tilde H n }\norm{\cY}\uas^2}{B} + \frac{\tilde H\uas^2}{2}\\
    \nonumber
    &+  \frac{\sqrt{\tilde H}B}{\sqrt{n}\norm{\cY}}\mathbb{E}[\hat L(w^*;S)] \\
    \nonumber
    & \leq \frac{\sqrt{\tilde H}B}{\sqrt{n}\norm{\cY}}\br{\mathbb{E}[\hat L(\hat w;S) - \hat L(w^*;S)]} + 
    \br{\frac{\sqrt{\tilde H n}\ybound}{B}+ \frac{\tilde H}{2}}\uas^2
    \\&+  \frac{\sqrt{\tilde H}B\ybound}{\sqrt{n}}
\end{align}
where the third inequality follows from AM-GM inequality, and last follows since $w^*$ is the optimal solution: $\mathbb{E}[\hat L( w^*; S)] = L(w^*; \cD) \leq L(0;\cD) \leq \ybound^2$. Finally, using the  lower bound on $n$, we have $\frac{\tilde H}{2} \leq \frac{\sqrt{\tilde H n}\ybound}{B}$, substituting which gives the claimed bound.
\end{proof}

\begin{lemma}
\label{lem:gd-average-stability}
The average argument stability for noisy GD (Algorithm \ref{alg:noisySGD}) run for $T$ iterations with step size $\eta \leq \frac{4}{\tilde H}$ is bounded as 
 $$\uas^2 \leq \frac{8\tilde H \eta^2 T}{n} \frac{1}{n}\sum_{j=1}^T \mathbb{E}[\eloss(w_j;S) - \eloss(w^*;S)] + \frac{8\tilde H \eta^2 T\ybound^2}{n}.$$
\end{lemma}
\begin{proof}

The uniform argument stability analysis for (Noisy) (S)GD is limited to the Lipschitz setting \cite{HRS15, bassily2019private,bassily2020stability} and therefore not directly applicable.
We therefore need to modify the arguments to give an average stability analysis in smooth (non-Lipschitz) case.

Let $\hat w:=\cA(S)$,
 $S^{(i)}$ be the dataset where the $i$-th data point is replaced by an i.i.d. point $(x',y')$ and let $\hat w^{(i)}$ be the corresponding output of $\cA$. Moroever, let $w_i$ denote the iterate of noisy SGD on dataset $S$ and similarly $\tilde w_i^{(i)}$ for dataset $S^{(i)}$.
 
 We simply couple the Gaussian noise sampled at each iteration to be equal on both datasets. Using the fact the the updates are non-expansive, we have
\begin{align*}
   \norm{w_{t+1}-w_{t+1}'}  &\leq \norm{w_t-w_t'} +  \frac{\eta\br{\norm{\nabla\ell(w_t;(x_i,y_i))} + \norm{\nabla\ell(w_t';(x';y'))}}}{n} \\
   & \leq \frac{\eta \sum_{j=1}^t \br{\norm{\nabla\ell(w_j;(x_i,y_i))} + \norm{\nabla\ell(w_j';(x';y'))}}}{n}.
\end{align*}

From the self-bounding property (Fact \ref{fact:self-boundedness}), $\br{\norm{\nabla\ell(w_j;(x_i,y_i))}} \leq 2\sqrt{\tilde H \ell(w_j;(x_i,y_i))}$). Therefore, we get,
\begin{align*}
    \mathbb{E}[\norm{w_{t}-w_{t}'}^2] &\leq \frac{4\tilde H\eta^2T}{n^2}\sum_{j=1}^t\br{\mathbb{E}[\ell(w_j;(x_i,y_i)) + \ell(w_j';(x',y'))]} \\
    & = \frac{8\tilde H \eta^2 T}{n^2}\sum_{j=1}^t \mathbb{E}[\eloss(w_j;S)].
\end{align*}
 
 For the average iterate,
 \begin{align*}
     \mathbb{E}\left[\norm{\hat w - \hat w^{(i)}}^2\right] &\leq \frac{1}{T^2}T\sum_{t=1}^T\mathbb{E}[\norm{w_{t}-w_{t}'}^2] \leq \frac{8\tilde H \eta^2 T}{n^2}\sum_{j=1}^T \mathbb{E}[\eloss(w_j;S)] \\
     & \leq \frac{8\tilde H \eta^2 T}{n} \frac{1}{n}\sum_{j=1}^T \mathbb{E}[\eloss(w_j;S) - \eloss(w^*;S)] + \frac{8\tilde H \eta^2 T}{n}\mathbb{E}[\eloss(w^*;S)] \\
     & \leq  \frac{8\tilde H \eta^2 T}{n} \frac{1}{n}\sum_{j=1}^T \mathbb{E}[\eloss(w_j;S) - \eloss(w^*;S)] + \frac{8\tilde H \eta^2 T\ybound^2}{n}
 \end{align*}
 where the last inequality follows since $w^*$ is the population risk minimizer: $\mathbb{E}[\hat L( w^*; S)] = L(w^*; \cD)] \leq L(0;\cD) \leq \ybound^2$.
\end{proof}

\begin{lemma}
\label{lem:noisy-sgd-erm}
Let $n\geq \frac{\tilde HB^2}{\ybound^2}$, $\eta = \min\br{\frac{B}{\sqrt{T}\max\br{\sqrt{\tilde H}\ybound, \sigma\sqrt{d}}}, \frac{1}{4\tilde H}}$,  $\sigma^2 = \frac{8G^2T\log{1/\delta}}{n^2\epsilon^2}$ and 
$T=n$. We have,
\begin{align*}
     \mathbb{E}\left[\eloss(\hat w;S)-\eloss(w^*;S)\right] \leq \frac{1}{T}\sum_{j=1}^T\mathbb{E}\left[\eloss(w_j;S)-\eloss(w^*;S)\right]\leq O\br{\frac{\sqrt{H}B\ybound}{\sqrt{n}} + \frac{\sqrt{d}G\log{1/\delta}}{n\epsilon}}.
\end{align*}
where $w^*$ is the (minimum norm) population risk minimizer.
\end{lemma}
\begin{proof}
From standard analysis of (S)GD,

\begin{align*}
    \mathbb{E}\left[\norm{w_{t+1} - w^*}^2\right] &\leq \mathbb{E}\left[ \norm{w_t - \eta \br{\nabla \eloss(w_t;S) + \xi_t} - w^*}^2\right] \\
    & \leq \mathbb{E}\left[\norm{w_t-w^*}^2\right] + \eta^2\mathbb{E}\left[\norm{\nabla \eloss(w_t;S)}^2\right] + \eta^2\sigma^2d -2\eta\mathbb{E}\left[\hat L(w_t;S)-\hat L(w^*;S)\right] \\
    &\leq \mathbb{E}\left[\norm{w_t-w^*}^2\right] + 4\eta^2\tilde H \mathbb{E}[\hat L(w_t;S)] +\eta^2\sigma^2d -2\eta\mathbb{E}\left[\hat L(w_t;S)-\hat L(w^*;S)\right]
\end{align*}

where the last inequality follows from self-bounding property (Fact \ref{fact:self-boundedness}).
Rearranging, and using the fact that $\mathbb{E}[\hat L( w^*; S) = L(w^*; \cD)] \leq L(0;\cD) \leq \ybound^2$ we get,
\begin{align*}
    \br{1-2\eta \tilde H}   \mathbb{E}\left[\hat L(w_t;S)-\hat L(w^*;S)\right] \leq 
    \frac{\mathbb{E}\left[\norm{w_t-w^*}^2 -  \norm{w_{t+1} - w^*}^2\right]}{2\eta} + 2\eta\br{\tilde H \ybound^2 + \sigma^2d}
\end{align*}

From the choice of $\eta$,
we have $ \br{1-2\eta \tilde H} \geq \frac{1}{2}$. Averaging over $T$ iterations,
we get that the average regret is,
\begin{align*}
    \frac{1}{T}\sum_{j=1}^T\mathbb{E}\left[\eloss( w_j;S)-\eloss(w^*;S)\right] \leq \frac{B^2}{\eta T} + 4\eta\br{\tilde H \ybound^2 + \sigma^2d}.
\end{align*}

Setting $\eta = \min\br{\frac{B}{\sqrt{T}\max\br{\sqrt{\tilde H}\ybound, \sigma\sqrt{d}}}, \frac{1}{4\tilde H}}$, we get
\begin{align*}
        \frac{1}{T}\sum_{j=1}^T\mathbb{E}\left[\eloss( w_j;S)-\eloss(w^*;S)\right]
     = O\br{\frac{\sqrt{\tilde H}B\ybound}{\sqrt{T}}+\frac{B\sqrt{d}\sigma}{\sqrt{T}} + \frac{\tilde HB^2}{T}}.
\end{align*}
Finally, substituting  
$\sigma^2 = \frac{8G^2T\log{1/\delta}}{n^2\epsilon^2}$, $T=n$
and using the lower bound on $n$ gives the claimed bound on average regret. Applying convexity to lower bound average regret by excess empirical risk of $\hat w$ gives the same 
bound on excess empirical risk.
\end{proof}

\subsection{High Dimension}

\begin{proof}[Proof of Theorem \ref{thm:smooth-jl}]
Let $\alpha\leq 1$ be a parameter to be set later.
From the JL property with $k = O\br{\frac{\log{2n/\delta}}{\alpha^2}}$, with probability at least $1-\delta/2$, for all data points $x_i$, $\norm{\Phi x_i}\leq \br{1+\alpha} \norm{x_i} \leq 2 \norm{\cX}$, and $\norm{\Phi w^*}^2 \leq 2\norm{w^*}^2\leq 2B^2$.

Further, by Lemma \ref{lem:smooth_lip}, for any $w$ in the embedding space with $\norm{w}^2\leq B$, with probability at least $1-\delta/2$, the  loss
is $G$ Lipschitz where $G= 2\norm{\cY}\sqrt{H}\norm{\cX} + 2HB\norm{\cX}^2$.
The privacy guarantee now follows from the privacy of Noisy SGD and post-processing. 

For the utility guarantee, 
let $L(w;\Phi \cD)$ and $\hat L(w;\Phi S)$ denote population and empirical risk (resp) where test and training feature vectors (resp) are mapped using $\Phi$. 
The excess risk can be decomposed as:
\begin{align}
\label{eqn:jl-smooth-decomposition}
    \mathbb{E}\left[L(\Phi^\top \tilde w;\cD) - L(w^*; \cD)\right] = \mathbb{E}\left[L(\tilde w;\Phi\cD) - L(\Phi w^*;\Phi \cD)\right]  +\mathbb{E}\left[L(\Phi w^*;\Phi \cD) -  L(w^*;\cD )\right].
\end{align}
The first term in Eqn. \eqref{eqn:jl-smooth-decomposition} is bounded by the utility guarantee of the DP-SCO method. In particular, from Lemma \ref{lem:noisy-sgd-smooth-jl} (below),
we have

\begin{align*}
    \mathbb{E}[L(\tilde w;\Phi \cD) - L(\Phi w^*;\Phi\cD)] = O\br{
    \frac{\sqrt{H}\xbound B\ybound}{\sqrt{n}}+ 
    \frac{\br{\sqrt{ H}\xbound\ybound +  H\xbound^2 B^2}B\sqrt{k \log{1/\delta}}}{n\epsilon}}.
\end{align*}

The second term in Eqn. \eqref{eqn:jl-smooth-decomposition} is bounded by the JL property together with smoothness and the fact that $w^*$ is the optimal solution thus $\nabla L(w^*; \cD)=0$. 
This gives us
\begin{align*}
    \mathbb{E}[L(\Phi w^*; \Phi \cD)]- L(w^*;  \cD)] \leq \frac{H}{2}\mathbb{E}\left[\abs{\ip{\Phi w^*}{\Phi x} - \ip{w^*}{x}}^2\right] \leq \frac{\alpha^2 H\norm{w^*}^2\xbound^2}{2} \leq \tilde O\left(\frac{HB^2\xbound^2}{k}\right).
\end{align*}

Combining, we get,
\begin{align*}
\mathbb{E}\left[L(\Phi^\top \tilde w;\cD) - L(w^*; \cD)\right]\leq
   \tilde O\br{
    \frac{\sqrt{H}\xbound B\ybound}{\sqrt{n}}+ 
    \frac{\br{\sqrt{H}\xbound\ybound + H\xbound^2B}B\sqrt{k \log{1/\delta}}}{n\epsilon}+\frac{HB^2\xbound^2}{k}}.
\end{align*}

Setting $k = O\br{\frac{BH\norm{\cX}\log {2n/\delta} n\epsilon}{ G}}^{2/3}$ completes the proof.

\end{proof}

\begin{lemma}
\label{lem:noisy-sgd-smooth-jl}
Let $\Phi \in \bbR^{d \times k}$ be a data-oblivious JL matrix. 
Let $S=\bc{\br{x_i,y_i}}_{i=1}^n$ of $n$ i.i.d data points and let $\Phi S := \bc{(\Phi x_i, y_i)}_{i=1}^n$.
Let $\tilde w \in \bbR^k$ be the average iterate returned by Algorithm \ref{alg:noisySGD} with $\sigma^2 = O\br{\frac{G^2\norm{\cX}^2\log{1/\delta}}{n^2\epsilon^2}}$ on dataset $\Phi S$.
For $k=\Omega\br{\log{n}}$, the excess risk of $\tilde w$ on $\Phi \cD$ is bounded as,
\begin{align*}
    \mathbb{E}[L(\tilde w;\Phi \cD) - L(\Phi w^*;\Phi\cD)] \leq O\br{
    \frac{\sqrt{H}B\xbound\ybound}{\sqrt{n}}+ 
    \frac{\br{\sqrt{ H}\xbound\ybound +  H\xbound^2B}B\sqrt{k \log{1/\delta}}}{n\epsilon}}.
\end{align*}
\end{lemma}

\begin{proof}
Let $S^{(i)}$ be the dataset where the $i$-th data point is replaced by an i.i.d. point $(x',y')$ and let $\tilde w^{(i)}$ be the corresponding output of Noisy-SGD on $\Phi S^{(i)}$.
Define $\overline S := \bc{S,(x',y')}$
and let $H(\Phi,\overline S)$ denote an upper bound on the smoothness parameter of the family of loss function $\bc{\ell(w;(\Phi x;y))}_{(x,y)\in \overline S}$.

We want to apply Theorem \ref{thm:smooth-low-dim}, but the theorem requires that $n\geq \frac{2H\xbound^2B^2}{\ybound^2}$.
In the proof below, we will use it to bound $H(\Phi,\overline S)\leq \frac{n\ybound^2}{B^2}$.
We use the JL property to get this.
Let $\alpha \leq 1$ be a parameter to be set later.
Note that
$H(\Phi,\overline S) \leq H \sup_{(x,y)\in \overline S}\norm{\Phi x}^2 \leq H \br{1+\alpha}\norm{\cX}^2 \leq 2H\norm{\cX}^2$ with probability at least $1-\delta$ for $k = O\br{\frac{\log{2n/\delta}}{\alpha^2}}$.
Thus, if we assume $n\geq \frac{2H\xbound^2B^2}{\ybound^2}$, then w.h.p. $H(\Phi,\overline S)\leq \frac{n\ybound^2}{B^2}$.
Also from the JL property, $\norm{\Phi w^*}\leq 2B$.

Decomposing excess risk and writing generalization gap as on-average stability, we have
\begin{align*}
     \mathbb{E}_{\Phi,S}[L(\tilde w;\Phi\cD) - L(\Phi w^*;\Phi \cD)] &=   \mathbb{E}_{\Phi,S}[L(\Phi^\top\tilde w;\cD) - \hat L(\Phi^\top \tilde w;S)] + \mathbb{E}_{\Phi,S}[\hat L(\Phi^\top \tilde w;S)] - \hat L(\Phi w^*;\Phi S)] \\
     & = \mathbb{E}_{\overline S, \Phi}[\ell(\tilde w^{(i)}; (\Phi x_i,y_i)) - \ell(\tilde w; (\Phi x_i,y_i)) + \hat L(\Phi^\top \tilde w;S) - \hat L(\Phi w^*;\Phi S)].
\end{align*}

We now fix the randomness of 
$\overline S$ and bound the terms in high probability w.r.t.~the random $\Phi$.
Let $\ias = \norm{\hat w-\hat w^{(i)}}$ denote the instance argument stability.
Repeating the analysis in Theorem \ref{thm:smooth-low-dim}, from Eqn. \eqref{eqn:gen-gap-average-stability-eqn}, the first term is bounded as,
\begin{align}
\label{eqn:smooth-gd-jl-one}
\nonumber
    \ell(\tilde w^{(i)}; (\Phi x_i,y_i)) - \ell(\tilde w; (\Phi x_i,y_i)) &\leq \frac{\sqrt{H(\Phi,\overline S)} B}{\sqrt{n}Y}\br{\ell(\hat w;(\Phi x_i,y_i)) - \hat L(\Phi w^*;\Phi S)}  \\
    \nonumber
    &+ \br{\frac{\sqrt{H(\Phi,\overline S) n}\ybound}{B}+ \frac{H(\Phi,\overline S)}{2}}\ias^2 + \frac{\sqrt{H(\Phi,\overline S)B\ybound}}{\sqrt{n}}\\
    \nonumber
    &\leq \frac{2\sqrt{H}\xbound B}{\sqrt{n}Y}\br{\ell(\hat w;(\Phi x_i,y_i)) - \hat L(\Phi w^*;\Phi S)}  \\&+ \frac{4\sqrt{H n}\xbound\ybound}{B}\ias^2 + \frac{2\sqrt{HB\ybound}\xbound}{\sqrt{n}}
\end{align}
where the last inequality holds from application of JL property: with probability at least $1-\delta$, $H(\Phi,\overline S)\leq \frac{n\ybound^2}{B^2}$ (from the lower bound on $n$) and $H(\Phi,\overline S) \leq 2H\xbound^2$. 

As in the proof of Lemma \ref{lem:gd-average-stability}, $\ias$ is bounded as, 
\begin{align*}
    &\ias^2 \\&\leq \frac{8 H(\Phi, \overline S)  \eta^2 T}{n} \frac{1}{n}\sum_{j=1}^T \mathbb{E}[\ell(w_j;(\Phi x_i,y_i)) + \ell(w_j';(\Phi x',y')) - 2\eloss(\Phi w^*;\Phi S)] + \frac{8H(\Phi, \overline S)  \eta^2 T\ybound^2}{n} \\
    & \leq \frac{16 H\xbound^2 \eta^2 T}{n} \frac{1}{n}\sum_{j=1}^T \mathbb{E}[\ell(w_j;(\Phi x_i,y_i)) + \ell(w_j';(\Phi x',y')) - 2\eloss(\Phi w^*;\Phi S)] + \frac{16H\xbound^2 \eta^2 T\ybound^2}{n}.
\end{align*}

Substituting the above in equation \ref{eqn:smooth-gd-jl-one}, taking expectation with respect to $\overline S$ and from manipulations as in the proof of Theorem \ref{thm:smooth-low-dim}, we get that with probability at least $1-\delta$, we have 

\begin{align*}
    \mathbb{E}_{S}[L(\Phi^\top\tilde w;\cD) - \hat L(\Phi^\top \tilde w;S)] &\leq 
      2\br{\mathbb{E}[\hat L(\hat w;\Phi S) - \hat L(\Phi w^*;\Phi S)]} \\
      &+
 \frac{32}{n}\br{\sum_{j=1}^T \mathbb{E}[\eloss(w_j;\Phi S) - \eloss(w^*;\Phi S)]}
    +  \frac{34\sqrt{\tilde H}B\ybound}{\sqrt{n}}.
\end{align*}

Let $G(\Phi,S) = G=2\norm{\cY}\sqrt{H(\Phi, \overline S)} + 2 H(\Phi, \overline S)\norm{\Phi w^*}$ denote the Lispchitzness parameter of the family of loss functions $\bc{\ell(w;(\Phi x,y)}_{(x,y)\in \overline S}$.
From the analysis in Lemma \ref{lem:noisy-sgd-erm}, the average regret and excess empirical risk terms are both bounded by the following quantity with high probability.
\begin{align*}
     O\br{\frac{\sqrt{H(\Phi, \overline S)}B\ybound}{\sqrt{n}} + \frac{\sqrt{k}G(\Phi,S)\log{1/\delta}}{n\epsilon}} \leq 
    O\br{\frac{\sqrt{H}B\ybound}{\sqrt{n}} + \frac{\sqrt{k}G\log{1/\delta}}{n\epsilon}}.
\end{align*}

This gives the following high probability bound,
\begin{align*}
    \mathbb{E}[L(\tilde w;\Phi \cD) - L(\Phi w^*;\Phi\cD)] \leq O\br{
    \frac{\sqrt{H}B\xbound\ybound}{\sqrt{n}}+ 
    \frac{\br{\sqrt{ H}\xbound\ybound +  H\xbound^2B}B\sqrt{k \log{1/\delta}}}{n\epsilon}}.
\end{align*}

For the in-expectation bound, note that in the above proof the JL property was used for:
with probability at least $1-\delta$, $\sup_{(x,y)\in \overline S} \norm{\Phi x} \leq \br{1+\frac{\sqrt{\log{n/\delta}}}{k}}\norm{\cX}$ and $\norm{\Phi w^*}\leq \br{1+\frac{\sqrt{\log{n/\delta}}}{k}}B$.
All these quantities appear in the numerator in the above bound.
Therefore the excess risk random variable w.r.t $\Phi$ has a tail
with a $\operatorname{poly}\br{\frac{\sqrt{\log{n/\delta}}}{k}}$ term.
This is a (non-centered) sub-Weibull random variable and from equivalence of tail and moments bounds (e.g.~Theorem 3.1 in \cite{vladimirova2020sub}) 
and $\frac{\log{n}}{k}\leq O(1)$, we get the claimed expectation bound.

\end{proof}

\subsection{Constrained Regularized ERM with Output Perturbation} 
We here
state a key result from \cite{SST10}. 
Let $\mathfrak{R}_n(B,\norm{\cX})$ denote the expected Rademacher complexity of 
linear predictors with norm bound by $B$, with $n$ datapoints  and norm of each point bounded by $\norm{\cX}$.

\begin{theorem} \label{thm:smooth_rad} [\cite{SST10}]
Let $\ell$ be an $H$-smooth GLM and let $R$ be a bound on the loss function. For a dataset $S$ of $n$ i.i.d samples, with probability at least $1-\beta$, for all $w$ such that $\norm{w}\leq B$, we have
\begin{align}
\label{eqn:sst10-rademacher-complexity}
\nonumber
    L(w;\cD) \leq \hat L(w;S) +& O\Bigg(\sqrt{\hat L(w;S)}\br{ \sqrt{H}\operatorname{log}^{1.5}(n) \mathfrak{R}_n(B,\norm{\cX}) + \sqrt{\frac{R\log{1/\beta}}{n}}} \\&+  H\operatorname{log}^3(n)\mathfrak{R}_n^2(B,\norm{\cX})  + \frac{R\log{1/\beta}}{n}\Bigg).
\end{align}
\end{theorem}
\begin{corollary} \label{cor:smooth-reg-rad}

Let $\tilde{w} = \argmin\limits_{w\in\ball{B}}\bc{\hat L(w;S)}$ and 
$n\geq \frac{HB^2\xbound^2}{\ybound^2}$.
Then with probability at least $1-\beta$ under the randomness of $S$ we have
\begin{equation*}
    L(\tilde{w};\cD) - \hat L(\tilde{w};S) = \tilde O\br{\frac{\sqrt{H}B\xbound\ybound\sqrt{\log{1/\beta}}}{\sqrt{n}} + \frac{\ybound^2\log{1/\beta}}{\sqrt{n}}}.
\end{equation*}
Further, in expectation under the randomness of S it holds that
\begin{equation*}
    \mathbb{E}[L( \tilde w;\cD)- \hat L(\tilde w;S)] = \tilde O\br{\frac{ \sqrt{H}\ybound B\norm{\cX}}{\sqrt{n}} + \frac{\norm{\cY}^2}{\sqrt{n}}}.
\end{equation*}
\end{corollary}
\begin{proof}
In our application, $w$ will be the output of regularized ERM $\tilde w$, thus $\hat L(\tilde w;S) \leq \hat L_\lambda(0;S) \leq \norm{\cY}^2$.

From Lemma \ref{lem:smooth_loss_bound} we have that $R \leq 3(\ybound^2 + HB^2\xbound^2)$.
Also,
$\mathfrak{R}_n(B,\norm{\cX}) \leq O\br{\frac{B\norm{\cX}}{\sqrt{n}}}$ \cite{shalev2014understanding}.
We now plug in the quantities into the above theorem to get that with probability at least $1-\beta$,
\begin{align*}
     &L( \tilde w;\cD)- \hat L(\tilde w;S) \\&= \tilde O \Bigg(\frac{\sqrt{H}B\xbound}{\sqrt{n}}\br{\ybound+\frac{\sqrt{H}B\xbound}{\sqrt{n}}} \\&+  \frac{\br{\ybound+\sqrt{H}B\xbound}\sqrt{\log{1/\beta}}}{\sqrt{n}}\br{\ybound+ \frac{\br{\ybound+\sqrt{H}B\xbound}\sqrt{\log{1/\beta}}}{\sqrt{n}}}\Bigg)\\
     &=\tilde O\br{\frac{\sqrt{H}B\xbound\ybound\sqrt{\log{1/\beta}}}{\sqrt{n}} + \frac{\ybound^2\log{1/\beta}}{\sqrt{n}}}
\end{align*}
where the last follows by simplifications using the lower bound on $n$.

For the in expectation result, observe that the above Eqn \eqref{eqn:sst10-rademacher-complexity} is a tail bound for a sub-Gamma random variable with variance parameter 
$O\br{\frac{\sqrt{H}B\xbound\ybound}{\sqrt{n}}}$ and scale parameter $O\br{\frac{\ybound^2}{\sqrt{n}}}$.
From the equivalence of tail and moment bounds of sub-Gamma random variables,
\cite{boucheron2013concentration}
we get the claimed in-expectation bound.
\end{proof}

\subsection{Proof of Theorem \ref{thm:smooth-output-perturbation}}
\begin{proof}
Recall from Lemma \ref{lem:smooth_lip} that the (un-regularized) loss is $G$-Lipschitz on the constraint set with $\norm{w}\leq B$, where $\tilde G:=\br{\sqrt{H}\norm{\cY}+HB\norm{\cX}}{\norm{\cX}}$. 
Note that from a standard analysis \cite{Bousquet:2002}, we get that $\ell_2$ sensitivity (or uniform argument stability) is $O\br{\frac{G}{n\lambda}}$, hence $\sigma^2 = O\br{\frac{G^2\log{1/\delta}}{\lambda^2n^2\epsilon^2}}$ ensures $(\epsilon, \delta)$-DP.

For the utility analysis, the excess risk can be decomposed as follows,
\begin{align*}
 \mathbb{E}[L(\hat w;\cD)-L(w^*;\cD)] &=
 \mathbb{E}[L(\tilde w;\cD)- \hat L(\tilde w;S)] + \mathbb{E}[\hat L(\tilde w;S)-\hat L(w^*;S)]+\mathbb{E}[L(\tilde w+\xi;\cD) - L(\tilde w;\cD)].
\end{align*}

The first term is bounded by the Rademacher complexity result (Corollary \ref{cor:smooth-reg-rad}).

The second term is simply bounded by $\frac{\lambda}{2}\norm{\tilde w}^2 \leq \frac{\lambda}{2}B^2$ since $\tilde w$ lies in the constraint set.
The third term is bounded using smoothness as follows:

\begin{align*}
    \mathbb{E}[L(\tilde w+\xi;\cD) - L(\tilde w;\cD)] &= \mathbb{E}\left[\phi_y\br{\ip{\tilde w + \xi}{x}}-\phi_y\br{\ip{\tilde w}{x}}\right] \\&\leq \mathbb{E}\left[\phi_y'(\ip{\tilde w}{x}) \ip{\xi}{x} + \frac{H}{2}\abs{\ip{\xi}{x}}^2\right] \\
    & \leq \frac{H}{2}\sigma^2\norm{\cX}^2 = O\br{\frac{HG^2 \norm{\cX}^2\log{1/\delta}}{\lambda^2 n^2\epsilon^2}}
\end{align*}
where the last inequality follows since $\mathbb{E}\xi = 0$ and $\ip{\xi}{x}\sim \cN(0,\sigma^2\norm{x}^2)$.
We now plug in $\lambda = \br{\frac{G\sqrt{H}\norm{\cX}}{B}}^{2/3}\frac{ \br{\log{1/\delta}}^{1/3}}{\br{n\epsilon}^{2/3}}$.
and $G=\br{\sqrt{H}\norm{\cY}+HB\norm{\cX}}{\norm{\cX}}$ to get,

\begin{align*}
    \mathbb{E}[L(\hat w;\cD)-L(w^*;\cD)] &\leq  \tilde O\Bigg(\frac{ \sqrt{H}B\norm{\cX}\norm{\cY} + \ybound^2}{\sqrt{n}} 
    +\frac{\br{\sqrt{H}B\norm{\cX}}^{4/3}\norm{\cY}^{2/3}+ \br{\sqrt{H}B\norm{\cX}}^2}{(n\epsilon)^{2/3}}\Bigg).
\end{align*}
\end{proof}

\subsection{Proof of Theorem \ref{thm:smooth-lower-bound}} \label{app:smooth-lowerbound}
\begin{proof}
Define $F(w;S) = \frac{1}{n}\sum_{(x,y)\in S} (\ip{w}{x}-y)^2$.
Let $d'<\min\bc{n,\dimension}$
and $b,p \in [0,1]$ 
be parameters to be chosen later. For any  $\bssigma \in \bc{\pm 1}^{d'}$, define the dataset $S_{\bssigma}$ which consists of the union of $d'$ subdatasets, $S_1,...,S_{d'}$ given as follows. Set $\frac{pn}{d'}$ of the feature vectors in $S_{j}$ 
as $\scale e_j$ (the rescaled $j$'th standard basis vector) and the rest as the zero vector. Set $\frac{pn}{2d}(1+b)$ of the labels as $\bssigma_j \ybound$ and $\frac{pn}{2d}(1-b)$ labels as $-\bssigma_j \ybound$. 
Let $w^{\bssigma} = \argmin_{w\in\re^d}\bc{F(w;S_{\bssigma})}$ be the ERM minimizer of $F(\cdot;S_{\bssigma})$.
Following from Lemma 2 of \cite{shamir_2015} we have that for any $\bar{w} \in \re^{d}$ that
\begin{equation} \label{eq:dim_loss_bound}
F(\bar{w};S_{\bssigma}) - F(w^{\bssigma};S_{\bssigma}) \geq \frac{p\scale^2}{2d'}\sum_{j=1}^{d'}{(\bar{w_j} - w^{\bssigma}_j)^2}. 
\end{equation} 

We will now show lower bounds on the per-coordinate error. 
Consider any $\bssigma$ and $\bssigma'$ which
differ only at index $j$ for some $j\in[d']$.
Note that the datasets $S_{\bssigma}$ and $S_{\bssigma'}$  differ in $\Delta=\frac{pn}{2d'}[(1+b)-(1-b)] = \frac{pbn}{d'}$ points.
Let $\tau = w_j^{\bssigma} = \frac{\ybound b}{\xbound}$ and $\tau' = w_j^{\bssigma'} = -\frac{\ybound b}{\xbound}$ (i.e. the $j$ components of the empirical minimizers for $S$ and $S'_j$ respectively).
Note that $|w_j^{\bssigma} - w_j^{\bssigma'}| = \frac{2\ybound b}{\xbound}$. 
We thus have by Lemma \ref{lem:ch} that for a certain $b=b(\epsilon,n,d,B,p,\ybound)$, $\cA$ must satisfy 
\begin{equation} \label{eq:ybaH_bound}
\ex{}{|\cA(S_{\bssigma})_j - w_j^{\bssigma}| + |\cA(S_{\bssigma'})_j - w_j^{\bssigma'}|} \geq \frac{1}{4}\frac{\ybound b}{\xbound}. 
\end{equation}
Since we need $\Delta \leq \frac{1}{\epsilon}$, we must set $b \leq \frac{d'}{pn\epsilon}$.
Furthermore, if we are interested in problems with minimizer norm at most $B$, we need $b \leq \frac{\scale B}{\ybound \sqrt{d'}}$ to ensure the norm of the minimizer is bounded by $B$. 
Balancing these two restrictions, we set 
$d' = \big(\frac{p\scale Bn\epsilon}{\ybound}\big)^{2/3}$ which yields 
$b = \big(\frac{\scale B}{\ybound \sqrt{pn \epsilon}}\big)^{2/3}.$
Assuming such settings of $b$ and $d'$ are possible (e.g. $b\in[0,1]$) 
we can apply Jensen's inequality and the fact that $(a+b)^2 \leq 2(a^2+b^2)$ to Eqn. \eqref{eq:ybaH_bound} to obtain

\begin{align*}
    \ex{}{|\cA(S_{\bssigma})_j - w_j^{\bssigma}|^2 + |\cA(S_{\bssigma'})_j - w_j^{\bssigma'}|^2} \geq \frac{B^{4/3}\ybound^{2/3}}{32(\scale pn\epsilon)^{2/3}}.
\end{align*}
We will now show this implies there exists a $\bssigma\in \bc{\pm 1}^{d'}$ such that $F(\cA(S_{\bssigma})) - F(w^{\bssigma}) = \Omega(\frac{(\scale B)^{4/3}\ybound^{2/3}p^{1/3}}{(n\epsilon)^{2/3}})$. To prove this, we have the following analysis.
Let $U = \bc{\pm 1}^{d'}$ and let $\bssigma_{-j}$ denote the vector $\bssigma$ with its $j$'th component negated. 
We have 
\begin{align*}
\sup_{\bssigma \in U}\bc{\ex{}{F(\cA(S_{\bssigma})) - F(w^{\bssigma})}} 
&\geq \frac{1}{|U|}\sum_{\bssigma\in U} \ex{}{F(\cA(S_{\bssigma}))-F(w^{\bssigma})} \\
&= \frac{p\scale^2}{d'|U|} \sum_{j\in[d']} \sum_{\bssigma\in U} \ex{}{|\cA(S_{\bssigma})_j - w_j^{\bssigma}|^2} \\
&= \frac{p\scale^2}{d'|U|} \sum_{j\in[d']} \sum_{\bssigma\in U: \bssigma_j = 1} \mathbb{E}\big[|\cA(S_{\bssigma})_j - w_j^{\bssigma}|^2  
+ |\cA(S_{\bssigma_{-j}})_j - w_j^{\bssigma_{-j}}|^2\big] \\
&\geq \frac{(\scale B)^{4/3}\ybound^{2/3}p^{1/3}}{128 (n\epsilon)^{2/3}}. 
\end{align*}
We recall this bound holds providing the settings of $d'$ and $b$ fall into the range $[1,\min\bc{n,\dimension}]$ and $[0,1]$ respectively.
First note $b>0$ always. 
Furthermore,
$d' < \min\bc{n,\dimension}$ and $b<1$ whenever 
\begin{equation} \label{eq:param_condition}
\frac{\scale^2B^2}{\ybound^2n\epsilon} \leq p \leq \min\bc{1,\frac{\dimension^{3/2}\ybound}{\scale Bn\epsilon}}.  
\end{equation}
 
In the following, assume $B \leq \frac{\ybound\min\bc{\sqrt{n\epsilon},\sqrt{\dimension}}}{\xbound}$.
Note this is no loss of generality as we can obtain problem instances with arbitrarily large $B$ by adding a dummy point with $x=c'e_{d'+1}$ and $y=\ybound$ where $c'$ is arbitrarily small so that the minimizer norm is $B$.
Under this assumption on $B$, using the restrictions on $p$, it can be verified that $d' > 1$ whenever $\epsilon > \frac{1}{n}$. Thus we have $b\in[0,1]$ and $d'\in [1,\min\bc{n,\dimension}]$ as required.
Furthermore this assumption on $B$ implies
$\frac{\scale^2B^2}{\ybound^2n\epsilon} \leq \min\bc{1,\frac{\dimension^{3/2}\ybound}{\scale Bn\epsilon}}$
and thus a valid setting of $p$ is possible.
We now turn to setting $p$ in a way which satisfies \eqref{eq:param_condition}. We consider two cases, the high and low dimensional regimes.
\newline\textbf{Case 1:} $\dimension \geq \br{\frac{B\scale n\epsilon}{\ybound}}^{2/3}$.
Setting $p=1$ gives a lower bound of $\Omega\br{\min\bc{\ybound^2,\frac{B^{4/3}\scale^{4/3}\ybound^{2/3}}{32(n\epsilon)^{2/3}}}}$, where the min with $\ybound^2$ from the upper bound on $B$. 
\newline\textbf{Case 2:} $\dimension \leq \br{\frac{B\scale n\epsilon}{\ybound}}^{2/3}.$
Setting $p=\frac{d^{3/2}\ybound}{B\scale n\epsilon}$ we obtain a bound of $\Omega\br{\min\bc{\ybound^2,\frac{\sqrt{\dimension}B\norm{\cX}\norm{\cY}}{n\epsilon}}}$ 
which we note is no larger than the bound from Case 1 in the low dimensional regime and no smaller than the bound from Case 1 in the high dimensional regime. Thus we can write the total bound as the minimum of these two bounds. 
The $\ybound^2$ term again comes from the restriction on $B$.

To obtain results for arbitrary $H$, we can set $F(w;S) = \frac{H}{2n}\sum_{(x,y)\in S} (\ip{w}{x}-\frac{2}{\sqrt{H}}y)^2$. This satisfies $H$-smoothness and loss bounded at zero by $\ybound^2$. Then substituting $\ybound$ in the previous expressions for $2\ybound/\sqrt{H}$ and multiplying through by $H/2$ one obtains the claimed bound. 
\end{proof}

\section{Missing Proofs from Section \ref{sec:lip_glm} (Lipschitz GLMs) } \label{app:lip_glm}
\subsection{Proof of Theorem \ref{thm:lipschitz-output-perturbation}} 
\begin{proof} 
\label{app:proof-lip-output-perturb}
Note that from a standard analysis \cite{Bousquet:2002}, we get that $\ell_2$ sensitivity of the regularized minimizer $\tilde{w}$ is $\br{\frac{2G\xbound}{n\lambda}}$, hence $\sigma^2 =\frac{4G^2\xbound^2\log{1/\delta}}{\lambda^2n^2\epsilon^2}$ ensures $(\epsilon, \delta)$-DP.

For the utility analysis, the excess risk can be decomposed as follows,
\begin{align*}
 \mathbb{E}[L(\hat w;\cD)-L(w^*;\cD)] &=
 \mathbb{E}[L(\tilde w;\cD)- \hat L(\tilde w;S)] + \mathbb{E}[\hat L(\tilde w;S)-\hat L(w^*;S)]+\mathbb{E}[L(\tilde w+\xi;\cD) - L(\tilde w;\cD)]
\end{align*}

The first term is bounded as $O\br{\frac{BG\xbound}{\sqrt{n}}}$ from Rademacher complexity results on bounded linear predictors \cite{shalev2014understanding}.

The second term is simply bounded by $\frac{\lambda}{2}\norm{\tilde w}^2 \leq \frac{\lambda}{2}B^2$ since $\tilde w$ is the regularized ERM. The third term is bounded via the following
\begin{align*}
    \mathbb{E}[L(\tilde w+\xi;\cD) - L(\tilde w;\cD)] &= \mathbb{E}\left[\phi_y\br{\ip{\tilde w + \xi}{x}}-\phi_y\br{\ip{\tilde w}{x}}\right] \\
    &\leq \mathbb{E}\left[G|\ip{\xi}{x}|\right] \\
    & \leq G\sigma\norm{\cX} \leq \frac{4G^2 \xbound^2\sqrt{\log{1/\delta}}}{\lambda n \epsilon}
\end{align*}
where the last inequality follows since $\mathbb{E}\xi = 0$ and $\ip{\xi}{x}\sim \cN(0,\sigma^2\norm{x}^2)$. Now plugging in $\lambda = \frac{G\xbound\br{\log{1/\delta}}^{1/4}}{B\sqrt{n\epsilon}}$ obtains the following result
\begin{align*}
    \mathbb{E}[L(\hat w;\cD)-L(w^*;\cD)] &= O\br{\frac{BG\xbound}{\sqrt{n}} + \frac{BG\xbound\log{1/\delta}^{1/4}}{\sqrt{n\epsilon}}}.
\end{align*}

\end{proof}

\subsection{Upper Bound using JL Method}
\begin{theorem}
\label{thm:lipschitz-jl}
Let $k = O\br{\log {2n/\delta} n\epsilon},
\sigma^2 = \frac{8TG^2\norm{\cX}^2\log{2/\delta}}{n^2\epsilon^2},
\eta = \frac{B}{G\xbound\br{1+\frac{\sqrt{k\log{2/\delta}}}{n\epsilon}}T^{3/4}}$ and $T=n^2$.
Algorithm \ref{alg:jl-constrained-dp-erm} satisfies $(\epsilon,\delta)$-differential privacy. 
Given a dataset $S$ of $n$ i.i.d samples, 
the excess risk of its output $\tilde w$ is bounded as
\begin{align*}
    \mathbb{E}[\excessrisk(\out)] 
= \tilde O\br{\frac{GB\norm{\cX}}{\sqrt{n}}+\frac{GB\xbound}{\sqrt{n\epsilon}}}.
\end{align*}
\end{theorem}
\label{app:proof-lip-jl}
\begin{proof}
Let $\alpha \leq 1$ be a parameter to be set later. From the JL property, with $k=O\br{\frac{\log{2n/\delta}}{\alpha^2}}$, with probability at least $1-\frac{\delta}{2}$, for feature vectors have $\norm{\Phi x_i}\leq \br{1+\alpha}\norm{x_i}\leq 2\norm{\cX}$, and $\norm{\Phi w^*}^2 \leq 2\norm{w^*}^2\leq 2B^2$.
Thus, gradient of loss for data point $(x,y)$ in $S$ at any $w$ is bounded as $\norm{\ell(w;(\Phi x,y)} = \phi_y'(\ip{w}{\Phi x})\norm{\Phi x} \leq 2G\norm{\cX}$. 
The privacy guarantee thus follows from the privacy of DP-SCO and post-processing.

For the utility guarantee, we decompose excess risk as:
\begin{align}
\label{eqn:jl-lispchitz-decomposition}
    \mathbb{E}\left[\ploss(\hat w;\cD) - \ploss(w^*;\cD) \right] &= \mathbb{E}\left[ \ploss(\tilde w;\Phi \cD) -  \ploss(\Phi w^*;\Phi S)\right] +
    \mathbb{E}\left[ \ploss(\Phi  w^*;\Phi S) - \ploss(w^*;\cD) \right].
\end{align}

The first term in Eqn. \eqref{eqn:jl-lispchitz-decomposition} is bounded by the utility guarantee of the DP-SCO method. In particular, from Lemma \ref{lem:noisy-sgd-lispchitz-jl} (below), we have

\begin{align*}
   \mathbb{E}\left[ \ploss(\tilde w;\Phi \cD) -  \ploss(\Phi w^*;\Phi S)\right] 
    \leq \frac{GB\norm{\cX}}{\sqrt{n}} + O\br{\frac{G\norm{\cX}B\sqrt{k}}{n\epsilon}}.
\end{align*}

For the second term in Eqn \eqref{eqn:jl-lispchitz-decomposition}, we use JL-property and Lipschitzness of GLM: 
\begin{align*}
    \mathbb{E}\left[ \ploss(\Phi  w^*;\Phi S) - \ploss(w^*;\cD) \right] &\leq G\mathbb{E}\left[\abs{\ip{\Phi x}{\Phi w^*}- \ip{x}{w^*}}\right] \\
    &\leq \alpha G\norm{\cX}\norm{w^*} \\&= O\br{\frac{G\norm{\cX}B\sqrt{\log{2n/\delta}}}{\sqrt{k}}}.
\end{align*}

Combining, we get, 
\begin{align*}
    \mathbb{E}\left[\ploss(\hat w;\cD) - \ploss(w^*;\cD) \right] \leq \tilde O\br{\frac{G\norm{\cX}B}{\sqrt{n}} +\frac{G\norm{\cX}B\sqrt{k}}{n\epsilon}+ \frac{G\norm{\cX}B}{\sqrt{k}}}.
\end{align*}

Balancing parameters by setting $k = \tilde O(n\epsilon)$ gives the claimed bound.
\end{proof}

\begin{lemma}
\label{lem:noisy-sgd-lispchitz-jl}
Let $\Phi \in \bbR^{d \times k}$ be a data-oblivious JL matrix. Let $S=\bc{\br{x_i,y_i}}_{i=1}^n$ of $n$ i.i.d data points and let $\Phi S := \bc{(\Phi x_i, y_i)}_{i=1}^n$.
Let $\tilde w \in \bbR^k$ be the average iterate returned noisy SGD procedure with Gaussian noise variance $\sigma^2 = O\br{\frac{G^2\norm{\cX}^2\log{1/\delta}}{n^2\epsilon^2}}$ on dataset $\Phi S$.
For $k=\Omega\br{\log{n}}$, the excess risk of $\tilde w$ on $\Phi \cD$ is bounded as,
\begin{align}
    \mathbb{E}_{\Phi,S}[L(\tilde w; \Phi \cD) - L(\Phi w^*; \Phi \cD)] \leq O\br{\frac{GB\xbound}{\sqrt{n}} + \frac{GB\xbound\sqrt{k\log{1/\delta}}}{n\epsilon}}.
\end{align}
\end{lemma}
\begin{proof}
The proof uses the analysis for excess risk bound for Noisy SGD in \cite{bassily2020stability} with the JL transform.   
Let $S^{(i)}$ be the dataset where the $i$-th data point is replaced by an i.i.d. point $(x',y')$ and let $\tilde w^{(i)}$ be the corresponding output of Noisy-SGD on $\Phi S^{(i)}$.
Define $\overline S := \bc{S,(x',y')}$
and let $G(\Phi,\overline S)$ denote an upper bound Lipschitzness parameter of the family of loss functions $\bc{\ell(w;(\Phi x;y))}_{(x,y)\in \overline S}$.

We decompose the excess risk as:
\begin{align*}
     \mathbb{E}_{\Phi,S}[L(\tilde w;\Phi\cD) - L(\Phi w^*;\Phi \cD)] =   \mathbb{E}_{\Phi,S}[L(\Phi^\top\tilde w;\cD) - \hat L(\Phi^\top \tilde w;S)] + \mathbb{E}_{\Phi,S}[\hat L(\Phi^\top \tilde w;S)] - \hat L(\Phi w^*;\Phi S)].
\end{align*}

A well-known fact (see \cite{shalev2014understanding}) is that generalization gap is equal to on-average-stability:
\begin{align*}
    \mathbb{E}_{\Phi,S}[L(\Phi^\top\tilde w;\cD) - \hat L(\Phi^\top\tilde w;S) ] &= \mathbb{E}_{\overline S, \Phi}[\ell( \Phi^\top \tilde w^{(i)}; (x_i,y_i)) - \ell(\Phi^\top\tilde w; (x_i,y_i))] \\
    & = \mathbb{E}_{\overline S, \Phi}[\ell(\tilde w^{(i)}; (\Phi x_i,y_i)) - \ell(\tilde w; (\Phi x_i,y_i))].
\end{align*}

From the analysis in Theorem 3.3 from \cite{bassily2020stability}, it follows that
\begin{align}
\label{eqn:jl-lispchitz-noisy-gd-one}
    \ell(\tilde w^{(i)}; (\Phi x_i,y_i)) - \ell(\tilde w; (\Phi x_i,y_i)) \leq G(\Phi;\overline S)\norm{\tilde w^{(i)} -\tilde w} \leq O\br{G(\Phi;\overline S)^2\br{\eta\sqrt{T}+\frac{\eta T}{n}}}.
\end{align}
where the last equality follows from the GLM structure of the loss function.

From analysis of Noisy-SGD \cite{bassily2014private}, the other term (excess empirical risk) can be bounded as 
\begin{align}
\label{eqn:jl-lispchitz-noisy-gd-two}
     \hat L( \tilde w;\Phi S)] - \hat L(\Phi w^*;\Phi S) \leq O\br{\eta\br{ G(\Phi;\overline S)^2+ \frac{kG(\Phi;\overline S)^2\log{1/\delta}}{n^2\epsilon^2}} + \frac{\norm{\Phi w^*}^2}{T\eta}}
\end{align}
We now take expectation with respect to $\Phi$. Note that $G(\Phi,\overline S) = \sup_{(x,y)\in \overline S} \sup_{w}\abs{g_x(\ip{w}{\Phi x})}\norm{\Phi x}$, where $g_x$ is the an element of the sub-differential of $\phi_x$ at $\ip{w}{x}$. By the Lipschitzness assumption $\abs{g_x(\ip{w}{\Phi x})} \leq G$ and from the JL property, 
with probability at least $1-\delta$, $\sup_{(x,y)\in \overline S} \norm{\Phi x} \leq \br{1+\frac{\sqrt{\log{n/\delta}}}{k}}\norm{\cX}$.
Similarly, $\norm{\Phi^\top w^*}\leq \br{1+\frac{\sqrt{\log{n/\delta}}}{k}}B$.
Observe that the tail is that of a (non-centered) sub-Gaussian random variable with variance parameter $O\br{\frac{1}{k}}$. Using the equivalence of tail and moment bounds of sub-Gaussian random variables \cite{boucheron2013concentration} and the fact that $\frac{\log{n}}{k}\leq O(1)$, we get the following bounds for Eqn. \eqref{eqn:jl-lispchitz-noisy-gd-one} and \eqref{eqn:jl-lispchitz-noisy-gd-two}:

\begin{align*}
    \mathbb{E}_{\Phi}\left[\ell(\tilde w^{(i)}; (\Phi x_i,y_i)) - \ell(\tilde w; (\Phi x_i,y_i)\right] \leq  O\br{G^2\norm{\cX}^2\br{\eta\sqrt{T}+\frac{\eta T}{n}}}
\end{align*}

\begin{align*}
    \mathbb{E}_\Phi\left[\hat L( \tilde w;\Phi S)] - \hat L(\Phi w^*;\Phi S)\right] \leq \tilde O\br{\eta\br{ G^2\norm{\cX}^2+ \frac{kG^2\norm{\cX}^2\log{1/\delta}}{n^2\epsilon^2}} + \frac{B^2}{T\eta}}.
    \end{align*}

Finally, as in \cite{bassily2020stability}, taking expectation with respect $\overline S$ in the two inequalities above, setting $\eta = \frac{B}{G\xbound\br{1+\frac{\sqrt{k \log{1/\delta}}}{n\epsilon}}T^{3/4}}$ and $T = n^2$ and combining gives the claimed bound.
\end{proof}

\subsection{Proof of Theorem \ref{thm:lipschitz_lower}}
The proof follows from the more general Theorem \ref{thm:lipschitz_lower_lp} stated below. Instantiating  Theorem \ref{thm:lipschitz_lower_lp} with $p=q=2$ satisfies all the requirements of our Theorem \ref{thm:lipschitz_lower}. Finally, let $w^*$ be the population risk minimizer of the hard instance in Theorem \ref{thm:lipschitz_lower_lp}. We then have,
\begin{align*}
     \mathbb{E}[L(\cA(S);\cD) - \min_{w} L(w;\cD)]
     &=  \mathbb{E}[L(\cA(S);\cD) -  L(w^*;\cD)]\\
    & \geq 
    \mathbb{E}[L(\cA(S);\cD) - \min_{w:\norm{w}_2\leq B}L(w;\cD)] \\
    \nonumber
    &\geq \mathbb{E}[L(\hat w;\cD) - L(\tilde \w;\cD)]\\
    &=\Omega\br{GB\xbound\min\br{1,\frac{1}{\sqrt{n\epsilon}},\frac{\sqrt{\rank}}{n\epsilon}}}.
\end{align*}
This completes the proof.

\subsection{Lower bound for \label{sec:lower-bound-lp}
Non-Euclidean DP-GLM}
\begin{theorem}
\label{thm:lipschitz_lower_lp}
Let $G,\norm{\cY},\norm{\cX}, B>0$, $\epsilon \leq 1.2, \delta \leq \epsilon$ and  $p,q\geq 1$. For any $(\epsilon,\delta)$-DP algorithm $\cA$, there exists 
sets $\cX$ and $\cY$
such that for any $x \in \cX$, $\norm{x}_q\leq 1$,
a
distribution $\cD$ over $\cX \times \cY$,
a $G$-Lipschitz GLM loss bounded at zero by $\norm{\cY}$ and a
$\tilde w$ with
$\norm{\tilde w}_p \leq B$ such that the output of $\cA$ on $S\sim\cD^n$ 
satisfies
\begin{align*}
    \mathbb{E}[L(\cA(S);\cD) - L(\tilde w;\cD)]
    =\Omega\br{GB\xbound\min\br{1,\frac{1}{\br{n\epsilon}^{1/p}},\frac{\br{\rank}^{(p-1)/p}}{n\epsilon}}}.
\end{align*}
\end{theorem}
\begin{proof}
The construction below is from \cite{SSTT21} with some changes.
We provide the complete proof below.
Consider the following $1$-Lipschitz loss function:
\begin{align*}
    \ell(w;(x,y)) = \abs{y-\ip{w}{x}}.
\end{align*}

Firstly, we argue that we can assume $\norm{\cY} = \infty$. This is because for any arbitrarily large value of $y$ we can translate the above function below so that the the loss is bounded by $\norm{\cY}$. However, this translation doesn't change the excess risk.
Also, as in \cite{bassily2014private}, it suffices to consider $G=1$, since we can simply scale the $1$-Lipschitz loss  function and get a factor of the $G$ in the lower bound.
Let $d' \leq \min\br{\rank,n\epsilon}$, $0\leq \alpha\leq 1$ and $\beta>0$ be parameters to be set later.
Since $d'\leq \rank$, without loss of generality, we will represent the features $x$ as $d'$ dimensional vectors. 

Consider a distribution $\cD$, where $x = e_0: = \vzero$ with probability $1-\alpha$, and with probability $\alpha$, $x \sim \operatorname{Unif}(\bc{
\norm{\cX}e_i}_{i=1}^{d'})$. 
Note that $\norm{x}_q \leq \norm{\cX}$ and for any $q\geq 1$ for $x \in \supp(\cD_x)$ where $\cD_x$ denotes the marginal of $\cD$ w.r.t. the $x$ variable.
This implies the loss $w\mapsto \ell(w;(x,y))$ is $\norm{\cX}$-Lipschitz in $\ell_q$-norm, when $x \in \supp(\cD_x)$ and for all $y$.
Let the fingerprinting code $z \in \bc{0,1}^{d'}$ be drawn  from a product distribution with mean $\mu \in [0,1]^{d'}$ where each co-ordinate $\mu_i \sim \operatorname{Beta}(\beta,\beta)$. 
Finally we have $y = \frac{B}{\br{d'}^{1/p}}\ip{x}{z}$.
Let $S=\bc{(x_i,y_i)}_{i=1}^n$ be $n$ i.i.d. samples drawn from $\cD$.
Define $\tilde w = \frac{B}{\br{d'}^{1/p}}\mu$; note that $\norm{\tilde w}_p \leq B$.

Let $\cA$ be any $(\epsilon,\delta)$-DP algorithm, which given $S$ outputs $\hat w$. Its excess risk with respect to $\tilde w$ can be lower bounded as,
\begin{align}
\nonumber
    \mathbb{E}[L(\hat w;\cD) - L(\tilde w;\cD)] &=
    \nonumber
    \mathbb{E}[\abs{y-\ip{\hat w}{x}} - \abs{y-\ip{\tilde w}{x}}] \\
    & \geq \mathbb{E}[\abs{\ip{\hat w-\tilde w}{x}} - 2 \abs{y-\ip{\tilde w}{x}}].
    \label{eqn:lower-bound-lispschitz-glm-risk}
\end{align}

The last term above is upper bounded as:
\begin{align*}
    \mathbb{E}[\abs{y-\ip{\tilde w}{x}}] 
    =\frac{B}{\br{{}d'}^{1/p}}\mathbb{E}[\abs{\ip{z}{x}-\ip{\mu}{x}}] 
    = \frac{B\norm{\cX}}{\br{{d'}}^{1/p}}\frac{\alpha}{{d'}} \sum_{i=1}^{d'} \mathbb{E}\abs{z_i-\mu_i}.  
\end{align*}
By direct  computation, 
\begin{align*}
    \mathbb{E}\abs{z_i-\mu_i} = \mathbb{E}\left[\abs{1-\mu_i}\mu_i + \abs{-\mu_i}\br{1-\mu_i}\right] %
    =2\mathbb{E}\mu_i\br{1-\mu_i} = \frac{2\beta}{1+2\beta}. 
\end{align*}
This gives us that 
\begin{align}
\label{eqn:excess-risk-term2}
    \mathbb{E}[\abs{y-\ip{\tilde w}{x}}] 
    \leq \frac{2\alpha\beta B\norm{\cX}}{\br{1+2\beta}\br{{d'}}^{(p+1)/p}}. 
\end{align}
The first term in the right hand side Inequality \eqref{eqn:lower-bound-lispschitz-glm-risk} is,
\begin{align}
\label{eqn:excess-risk-term1}
    \mathbb{E}[\abs{\ip{\hat w-\widetilde{\mu}}{x}} = \frac{B}{\br{{d'}}^{1/p}}\mathbb{E}\abs{\ip{\frac{\br{{d'}}^{1/p}}{B}\hat w - \mu}{x}} = \frac{B\norm{\cX} \alpha}{{\br{d'}^{(p+1)/p}}}\mathbb{E}\sum_{j=1}^{d'}\abs{\frac{\br{{d'}}^{1/p}}{B}\hat w_j - \mu_j}.
\end{align}

Define $v \in \bbR^{d'}$ with $v_i = \frac{{\br{d'}^{1/p}}\hat w_i}{B}$. Note that,
\begin{align*}
    \sum_{j=1}^{d'}\abs{\frac{{\br{d'}^{1/p}}}{B}\hat w_j - \mu_j}  = \norm{v - \mu}_1 \geq \norm{v - \mu}_2 \geq 
    \norm{\hat v - \mu}_2 \geq \norm{\hat v - \mu}_2^2
\end{align*}
where the first inequality above follows from relationship between $\ell_1$ and $\ell_2$ norm, the second follows from projection property and $\hat v$ denotes the projection
of $v$ onto $[0,1]^d$, 
and the last inequality follows from boundedness of coordinates of $\hat v -\mu$.

The key step now is application of the fingerprinting lemma (simplified below, see Lemma B.1 in \cite{SSTT21} for complete statement) from \cite{steinke2017tight}
which roughly speaking, ``relates the error to correlation''; for any $\hat v \in [0,1]^{d'}$, we have
\begin{align}
\label{eqn:fingerprinting}
    \mathbb{E}\norm{\hat v-\mu}_2^2 \geq \frac{{d'}}{4\br{1+2\beta}} - \frac{1}{\beta}\sum_{i=1}^n \mathbb{E}\ip{\hat v}{z_i-\mu}.
\end{align}

As in \cite{SSTT21}, the correlation simplifies as
\begin{align*}
   \mathbb{E}\ip{\hat v}{z_i-\mu} &= \sum_{j=0}^{d'}\mathbb{E}[\ip{\hat v}{z_i-\mu} | x_i=e_j]\mathbb{P}[x_i=\xbound e_j]
   =\frac{\alpha}{{d'}}\mathbb{E}[\hat v_j\br{z_i-\mu}_j | x_i=\xbound e_j]
\end{align*}
where the last equality follows since $\hat v$ only depends on the coordinate of $z_i$ for which $x_i=1$.
Now, let $\hat v^{\sim i}$ denotes the solution obtained if $z_i$ were replaced by an independent sample. 
From boundedness, $\norm{\hat v_j\br{z_i-\mu}_j}_{\infty}\leq 1$.
Using differential privacy, we have,
\begin{align*}
    \mathbb{E}[\hat v_j\br{z_i-\mu}_j | x_i=\xbound e_j] \leq \br{e^\epsilon-1} \mathbb{E}\left[\abs{\hat v_j^{\sim i}\br{z_i-\mu}_j} | x_i=\xbound e_j\right] + \delta \leq \br{e^\epsilon-1}+ \delta \leq 3\epsilon
\end{align*}

where the last inequality uses the assumption that  $\epsilon\leq 1.2$ and $\delta \leq \epsilon$. 
Plugging this in Eqn. \eqref{eqn:fingerprinting}, we get 
\begin{align*}
     \mathbb{E}\norm{\hat v-\mu}_2^2 \geq \frac{{d'}}{4\br{1+2\beta}} - 3 \alpha n\epsilon.
\end{align*}
Plugging the above in Eqn. \eqref{eqn:excess-risk-term1}, we get
\begin{align*}
     \mathbb{E}[\abs{\ip{\hat w-\tilde w}{x}} \geq \frac{ B\norm{\cX}\alpha }{{d'}^{(p+1)/p}}\br{\frac{{d'}}{4\br{1+2\beta}} - 3 \alpha n\epsilon}.
\end{align*}

Using the above, and the bound in Eqn. \eqref{eqn:lower-bound-lispschitz-glm-risk}, we get that,
\begin{align*}
     \mathbb{E}[L(\hat w;\cD) - L(\tilde w;\cD)] &\geq 
     \frac{ B\norm{\cX}\alpha}{{d'}^{(p+1)/p}}\br{\frac{{d'}}{4\br{1+2\beta}} - 3 \alpha n\epsilon} - \frac{2\alpha\beta B\norm{\cX}}{\br{1+2\beta}\br{{d'}}^{1/p}} \\
     & = \frac{B\norm{\cX}\alpha}{\br{1+2\beta}\br{{d'}}^{1/p}} \br{\frac{1}{4} - \frac{3 \alpha\br{1+2\beta} n\epsilon}{{d'}} - 2\beta }.
\end{align*}

Now, we set $\beta = \frac{1}{16}$. 
When $\rank \leq 48n\epsilon$ we set $d'=\rank$ and $\alpha = \min\br{\frac{{d'}}{48\br{1+2\beta}n\epsilon},1}$. The minimum term becomes $\frac{{d'}}{48\br{1+2\beta}n\epsilon}$ and obtain an excess risk lower bound of,
\begin{align*}
    \mathbb{E}[L(\hat w;\cD) - L(\tilde w;\cD)] \geq \frac{B\norm{\cX}\br{{d'}}^{(p-1)/p}}{1024n\epsilon}.
\end{align*}

On the other hand, when $\rank > 48n\epsilon$, set $\alpha=1$ and ${d'} = \lfloor 48 n\epsilon \rfloor$ (so that it is at least $1$) to get an excess risk lower bound of,
\begin{align*}
    \mathbb{E}[L(\hat w;\cD) - L(\tilde w;\cD)] \geq \min \br{\frac{B\norm{\cX}}{16\br{n\epsilon}^{1/p} },B\norm{\cX}}.
\end{align*}
Finally, note that
we can arbitrarily increase the rank of the construction beyond $48n\epsilon$ by adding datapoints with orthogonal feature vectors of small enough magnitude and arbitrary labels.
Combining the two lower bounds obtains the claimed bound.
\end{proof}

\begin{corollary}
\label{corr:lipschitz_dp_scpo_lower_lp}

Let $G, B>0$, $\epsilon \leq 1.2, \delta \leq \epsilon$ and  $p,q\geq 1$. 
Let $\cW \subset \bbR^d$ such that for any $w\in \cW$, $\norm{w}_p\leq B$.
For any $(\epsilon,\delta)$-DP algorithm $\cA$, there exists a set $\cZ$, a
distribution $\cD$ over $\cZ$ and a loss function $w\mapsto \ell(w;z)$, which is convex, $G$-Lipschitz w.r.t. $\ell_q$ norm for $w \in \cW$, for all $z \in \cZ$ such that the output of $\cA$ on $S\sim\cD^n$ (which may not lie in $\cW$), 
satisfies
\begin{align*}
    \mathbb{E}_{\cA,S}[\mathbb{E}_{z\sim \cD}\ell(\cA(S);z) - \min_{w\in \cW}\mathbb{E}_{z\sim \cD}\ell(w;z)]
    =\Omega\br{GB\min\br{1,\frac{1}{\br{n\epsilon}^{1/p}},\frac{d^{(p-1)/p}}{n\epsilon}}}.
\end{align*}
\end{corollary}

Using generalization properties of differential privacy, we get the same bound as above for excess empirical risk;  see Corollary B.4 in \cite{SSTT21} for details.

\section{Missing Details for Section \ref{sec:adapt} (Adapting to \texorpdfstring{$\mnorm$}{})}
\subsection{Generalized Exponential Mechanism} \label{app:gem}
\begin{theorem}
\cite{RS15}
Let $K>0$ and $S\in\cZ^n$. Let $q_1,...,q_K$ be functions s.t. for any adjacent datasets $S,S'$ it holds that $|q_j(S)-q_j(S')|\leq\gamma_j:\forall j\in[K]$. There exists an Algorithm, $\gem$, such that when given sensitivity-score pairs $(\gamma_1,q_1(S)),...,(\gamma_N,q_N(S))$, privacy parameter $\epsilon>0$ and confidence parameter $\beta>0$, outputs $j\in[N]$ such that with probability at least $1-\beta$ satisfies $q_j(S) \leq \min\limits_{j\in[N]}\bc{q_j(S) + \frac{4\gamma_j\log{N/\beta}}{\epsilon}}$.
\end{theorem}

\subsection{Proof of Theorem \ref{thm:adapt}}
Note that by assumptions on $\cA$, the process of generating $w_1,...,w_K$ is $(\epsilon/2,\delta/2)-DP$. Furthermore, by Assumption \ref{asm:loss_bound} with probability at least $\delta/2$ the sensitivity values passed to $\gem$ bound sensitivity. Thus by the privacy guarantees of $\gem$ and composition we have that the entire algorithm is $(\epsilon,\delta)$-DP.

We now prove accuracy. 
In order to do so, we first prove that with high probability every $\tilde{L}_j$ is an upper bound on the true population loss of $w_j$. Specifically, define $\tau_j = \frac{\Delta(B_j)\log{4K/\beta}}{n} + \sqrt{\frac{4\ybound^2\log{4K/\beta}}{n}}$ (i.e. the term added to each $L(w_j;S_2)$ in Algorithm \ref{alg:adapt}). 
Note it suffices to prove
\begin{equation} \label{eq:ploss-est-bound}
    \mathbb{P}\left[\exists j\in[K]: |L(w_j;S_2) - L(w_j;\cD)| \geq \tau_j \right] \leq \frac{\beta}{2}.
\end{equation}
Fix some $j\in[K]$. Note that the non-negativity of the loss implies that $\ell(w_B^*;(x,y)^2) \geq 0$. The excess risk assumption then implies that  $\ex{(x,y)\sim\cD}{\ell(w_j;(x,y))^2}\leq 4\ybound^2$, which in turn bounds the variance. Further, with probability at least $1-\frac{\beta}{4K}$ it holds that for all $(x,y)\in S_2$ that $\ell(w,(x,y)) \leq \Delta_0 + \Delta(B)$. Thus by Bernstein's inequality we have
\begin{equation*}
    \mathbb{P}\left[|L(w;S_2) - L(w;\cD)| \geq t \right] \leq \exp{-\frac{t^2n^2}{\Delta(B_j)tn + 4n\ybound^2}} + \frac{\beta}{4K}
\end{equation*}
Thus it suffices to set $t=\frac{\Delta(B_j)\log{4K/\beta}}{n} + \sqrt{\frac{4\ybound^2\log{4K/\beta}}{n}}$ to ensure $ \mathbb{P}\left[|L(w;S_2) - L(w;\cD)| \geq t \right] \leq \frac{\beta}{2K}$. 
Taking a union bound over all $j\in K$ establishes \eqref{eq:ploss-est-bound}. We now condition on this event for the rest of the proof.

Now consider the case where $j^*\neq 0$ and $\mnorm \leq 2^K$. Note that the unconstrained minimizer $w^*$ is the constrained minimizer with respect to any $\ball{r}$ for $r \geq \norm{w^*}$. 
With this in mind, let $j'=\min\limits_{j\in[K]}\bc{j:w^*\in\ball{2^j}}$ (i.e. the index of the smallest ball containing $w^*$).
In the following we condition on the event that $\forall j\in[K],j\geq j'$, the parameter vector $w_j$ satisfies excess population risk at most $\err(2^j)$. We note by Assumption \ref{asm:loss_bound} that this (in addition to the event given in \eqref{eq:ploss-est-bound}) happens with probability at least $1-\frac{3\beta}{4}$.
By the guarantees of $\gem$, with probability at least $1-\beta$ we (additionally) have
\begin{align*}
    L(w_{j^*};\cD) \leq L(w_{j^*};S_2) + \tau_{j^*} &\leq \min\limits_{j\in[K]}\bc{L(w_{j};S_2) + \tau_{j} + \frac{4\Delta(B_j)\log{4K/\beta}}{n\epsilon}} \\
    &\leq L(w_{j'};S_2) + \tau_{j'} + \frac{4\Delta(B_{j'})\log{4K/\beta}}{n\epsilon} \\
    &\leq L(w_{j'};\cD) + 2\tau_{j'} + \frac{4\Delta(B_{j'})\log{4K/\beta}}{n\epsilon}.
\end{align*} 

Since $2^{j'} \leq \max\bc{2\mnorm,1}$ we have 
\begin{align*}
    L(w_{j^*};\cD) - L(w^*;\cD) &\leq L(w_{j'};\cD)- L(w^*;\cD) + 2\tau_{j'} + \frac{4\Delta(B_{j'})\log{4K/\beta}}{n\epsilon} \\
    &\leq \err(2\max\bc{\mnorm,1}) + 2\tau_{j'} + \frac{4\Delta(\max\bc{2\mnorm,1})\log{4K/\beta}}{n\epsilon} \\
    &\leq \err(2\max\bc{\mnorm,1}) + \sqrt{\frac{4\ybound^2\log{4K/\beta}}{n}} + \frac{5\Delta(\max\bc{2\mnorm,1})\log{4K/\beta}}{n\epsilon} 
\end{align*}
where the second inequality comes from the fact the assumption that $\mnorm\leq\ybound^2$. 
Now note that by the assumption that $\err(2^K) \geq \ybound^2$, whenever $\mnorm \geq 2^K$ it holds that $\ybound^2 \leq \err(\mnorm)$. However since the sensitivity-score pair $(0,\ybound^2)$ is passed to $\gem$, the excess risk of the output is bounded by at most $\ybound^2$ by the guarantees of $\gem$). 

\subsection{Proof of Theorem \ref{thm:smooth-reg-pert-adapt}}
\label{app:pf-smooth-reg-pert-adapt}
Let $\out$ denote the output of the regularized output perturbation method with boosting and noise and privacy parameters $\epsilon'=\frac{\epsilon}{K}$ and 
$\delta'=\frac{\delta}{K}$. We have by Theorem \ref{thm:boosted-smooth-output-perturbation} that with probability at least $1-\frac{\beta}{4K}$ that 
\begin{align*}
    L(\out;\cD) - L(w^*;\cD) &= \tilde O\Bigg(\frac{\sqrt{H}B\norm{\cX}\norm{\cY} + \norm{\cY}^2}{\sqrt{n}} +\frac{\br{\br{\sqrt{H}B\norm{\cX}}^{4/3}\norm{\cY}^{2/3}+ \br{\sqrt{H}B\norm{\cX}}^2}}{(n\epsilon)^{2/3}} \\&+ \frac{\br{\ybound^2 + HB^2\xbound^2}}{n\epsilon} + \frac{\br{\ybound+\sqrt{ H}B\xbound}}{\sqrt{n}}
    \Bigg).
\end{align*}
Note that this is no smaller than $\ybound^2$ when $B= \Omega\br{\max\bc{\frac{\ybound\sqrt{n\epsilon}}{\xbound\sqrt{H}},\frac{\ybound^2(n\epsilon)^{2/3}}{\sqrt{H}\xbound^2}}}$, and thus it suffices to set 
$K=\Theta\br{\log{\max\bc{\frac{\ybound\sqrt{n}}{\xbound\sqrt{H}},\frac{\ybound^2(n\epsilon)^{2/3}}{\sqrt{H}\xbound^2}}}}$ to satisfy the condition of the Theorem statement.

Let $\sigma_j$ denote the level noise used for when the guess for $\mnorm$ is $B_j$. To establish Assumption \ref{asm:loss_bound}, by Lemma \ref{lem:reg-pert-sensitivity} we have that this assumption is satisfies with $\Delta(B) = \ybound^2 + H\xbound^2\sigma_j^2\log{K/\min\bc{\beta,\delta}}) + HB^2\xbound^2$. In particular, we note for the setting of $\sigma_j$ implied by Theorem \ref{thm:boosted-smooth-output-perturbation} and the setting of $K$ we have for all $j\in[K]$ that $H\xbound^2\sigma_j^2 = \tilde{O}(\ybound^2)$. Thus
$\Delta(B) = \tilde{O}\br{\ybound^2 + HB^2\xbound^2}$. The result then follows from Theorem $\ref{thm:adapt}$.

\subsection{Stability Results for Assumption \ref{asm:loss_bound}} \label{app:stab-asm}
\begin{lemma}
Algorithm \ref{alg:noisySGD} run with constraint set $\ball{B}$ satisfies Assumption \ref{asm:loss_bound} with $\Delta(B) = \ybound^2 + HB^2\xbound^2$.
\end{lemma}
The proof is straightforward using Lemma \ref{lem:smooth_loss_bound} (provided in the Appendix). For the output perturbation method, we can obtain similar guarantees. Here however, we must account for the fact that the output may not lie in the constraint set.
We also remark that the JL-based method (Algorithm \ref{alg:jl-constrained-dp-erm}) can also enjoy this same bound. However, in this case one must apply the norm adaptation method to the intermediate vector $\tilde{w}$, as $\Phi^{\top}\tilde{w}$ may have large norm.
\begin{lemma} \label{lem:reg-pert-sensitivity}
Algorithm \ref{alg:constrained-reg-erm-output-perturbation} run with parameter $B$ and $\sigma$ satisfies Assumption \ref{asm:loss_bound} with $\Delta(B) = \ybound^2 + H\xbound^2\sigma^2\log{K/\delta}) + HB^2\xbound^2$
\end{lemma}
\begin{proof}
Note that since $S$ and $S'$ differ in only one point, it suffices to show that for any $(x,y),(x',y')$ that
$\ell(\out;(x,y)) \leq \ybound^2 + HB^2\xbound^2 + H\xbound^2\sigma^2\log{K/\delta}$
and similarly for $\ell(\out,(x',y'))$.
Let $w\in \ball{B}$ and let $\out = w + b$ where $b \sim \cN(0,\mathbb{I}_d \sigma^2)$. 
We have by previous analysis
$\ell(\out;(x,y)) \leq \ybound^2 + HB^2\xbound^2 + H\ip{b}{x}^2$. Since $\ip{b}{x}$ is distributed as a zero mean Gaussian with variance at most $\xbound^2\sigma^2$, we have $\mathbb{P}[|\ip{b}{x}| \geq t] \leq \exp{\frac{-t^2}{\xbound^2\sigma^2}}.$
Setting $t = \xbound\sigma\log{K/\delta}$ we obtain
$\mathbb{P}[|\ip{b}{x}|^2 \geq \xbound^2\sigma^2\log{K/\delta}] \leq \delta/K$. Thus with probability at least $1-\delta/K$ it holds that
$\ell(\out;(x,y)) \leq \ybound^2 + HB^2\xbound^2 + H\xbound^2\sigma^2\log{K/\delta}$. 
\end{proof}

\section{Missing Details for Boosting}
\label{sec:appendix_boosting}

\begin{algorithm}[h]
\caption{Boosting}
\label{alg:boosting}
\begin{algorithmic}[1]
\REQUIRE Dataset $S$, loss function $\ell$, Algorithm $\cA$, $\tilde \sigma$
privacy parameters $\epsilon, \delta$
\STATE Split the dataset $S$ into equally sized chunks $\bc{S_i}_{i=1}^{m+1}$
\STATE For each $i \in [m+1]$, $\hat w_i = \cA\br{S_{i},\frac{\epsilon}{2},\delta}$
\STATE $i^* = \argmax_{i\in [m]}\br{-\hat L(\hat w_{i};S_{m+1}) + \operatorname{Lap}(0,\tilde \sigma)}$
\ENSURE{$\hat w_{i^*}$}
\end{algorithmic}
\end{algorithm}

We state the result of the boosting procedure in a general enough setup so as apply to our proposed algorithms.
This leads to additional conditions on the base algorithm since our proposed methods may not produce the output in the constrained set.

\begin{theorem}
\label{thm:boosting}
Let $\ell$ be a non-negative, $\tilde H$ smooth, convex loss function.
Let $\epsilon,\delta>0$.
Let $\cA:(S,\epsilon,\delta)\mapsto \cA(S,\epsilon,\delta)$ be an algorithm such that 
\begin{enumerate}
\item $\cA$ satisfies $(\epsilon, \delta)$-DP
\item For any fixed $S$, $\cA(S)$ is $\gamma^2$ sub-Gaussian \cite{vershynin2018high}:
$$\sup_{\norm{u}=1}\mathbb{E}\left[\exp{\ip{\cA(S)}{u}^2/\gamma^2}\right]\leq 2$$

\item For any fixed $S$, $\mathbb{P}_{(x,y)}[\ell(\cA(S);(x,y))>  \lbounddef]< \beta $
\item Given a dataset $S$ of $n$ i.i.d. points, $\mathbb{E}[L(\cA(S);\cD)-\min_{w\in \cB_B}L(w;\cD)]\leq \err\br{n,\epsilon,\gamma}$
\end{enumerate}

Let $\tilde \sigma^2 = \frac{4(\ybound^2 + \tilde H\tilde \gamma^2\xbound^2)}{n\epsilon}$ and  $n_0 = \frac{16\gamma^2\operatorname{log}^8\br{4/\beta}\tilde H}{\ybound^2}$.
\cref{alg:boosting} with Algorithm $\cA$ as input satisfies $(\epsilon,\delta)$-DP.
Given a  dataset $S$ of $n\geq n_0$ samples,  with probability at least $1-\beta$, 
the excess risk of its output $\hat w_{i^*}$ is bounded as,

\begin{align*}
 L(\hat w;\cD) - L(w^*;\cD) &\leq \tilde O\Bigg(\err\br{\frac{n}{4\log{4/\beta}},\frac{\epsilon}{2},\gamma} + \frac{2\Delta(\gamma,\beta/2)}{n\epsilon} +\frac{2\lboundtwo}{n}\\&+ \frac{32\gamma \sqrt{\tilde H}\ybound}{\sqrt{n}} +  \frac{16\ybound}{\sqrt{n}} +\frac{128\tilde H\gamma^2}{n}\Bigg).
\end{align*}
\end{theorem}

We first establish the following concentration bound for convex $\tilde H$ smooth non-negative functions.

\begin{lemma}
\label{lem:boosting-concentration}
Let $\ell$ be a convex $\tilde H$ smooth non-negative function.
 Let $S$ be a dataset of $n$ i.i.d. samples. 
 Let $w$ be a random variable which is $\gamma^2$ sub-Gaussian and independent of $S$ and let $\lbounddef$ be such that $\mathbb{P}_{(x,y)}[\ell(w;(x,y))>  \lbounddef]\leq \beta$. Then, with probability at least $1-\beta$,
    \begin{align*}
    \hat L(w;S) & \leq \br{1+T(n,\beta)}L(w;\cD) + U(n,\beta)\\
    L(w;\cD) &\leq \br{1+T(n,\beta)}\hat L(w;S)+V(n,\beta)
\end{align*}
where $T(n,\beta) := \frac{4\gamma \log{4/\beta}\sqrt{\tilde H}}{\ybound\sqrt{n}}$, $U(n,\beta):=\frac{4\gamma\log{4/\beta}\ybound\sqrt{\tilde H}}{\sqrt{n}}+ \frac{\ybound\sqrt{\log{2/\beta}}}{\sqrt{n}}$ and\\ $V(n,\beta) := \frac{4\gamma \log{4/\beta}\sqrt{\tilde H}\ybound}{\sqrt{n}} + \frac{2\lbound \log{2/\beta}}{n}+ \frac{\ybound\sqrt{\log{2/\beta}}}{\sqrt{n}} +\frac{48\tilde H\gamma^2\operatorname{log}^2(4/\beta)}{n}$.
\end{lemma}

\begin{proof}
With probability at least $1-\frac{\beta}{4}$, for each $(x,y)\in S, \ell(w;(x,y))\leq \lbound$. We condition on this event and apply Bernstein inequality to the random variable $L(w;\cD)-\hat L(w;S)$:
\begin{align*}
    \mathbb{P}\left[\abs{L(w;\cD)-\hat L(w;S)} > t\right] \leq \exp{-\frac{3nt^2}{6n\mathbb{E}[\br{L(w;\cD)-\hat L(w;S)}^2]+2\lbound t}}
\end{align*}
This gives us that 
\begin{align}
\label{eqn:boosting_bernstein}
    \abs{L(w;\cD)-\hat L(w;S)} \leq \frac{\lbound \log{2/\beta}}{n} +
    \sqrt{\mathbb{E}\br{L(w;\cD)-\hat L(w;S)}^2\log{2/\beta}}
\end{align}

The term $\mathbb{E}[\br{L(w;\cD)-\hat L(w;S)}^2 = \frac{1}{n}\mathbb{E}[\br{\ell(w;(x,y))-\mathbb{E}[\ell(w;(x,y)]}^2]\leq \frac{1}{n}\mathbb{E}[(\ell(w;(x,y)))^2]$.

Now, 
\begin{align*}
    \mathbb{E}[(\ell(w;(x,y)))^2] &\leq 2\mathbb{E}[(\ell(w;(x,y)) - \ell(0;(x,y))^2] + 2\mathbb{E}[(\ell(0;(x,y)))^2] \\
    & \leq 2 \mathbb{E}[\br{\ip{\nabla \ell(w;(x,y))}{w}}^2] + 2\ybound^2
\end{align*}

where the last step follows from convexity. We now use the fact that $w$ is $\boostsubg$-sub-Gaussian, therefore $\ip{\nabla \ell(w;(x,y))}{w} \leq \gamma\sqrt{\log{4/\beta}} \norm{\nabla \ell(w;(x,y))}$ with probability at least $1-\beta/4$.
We now use self-bounding property of non-negative smooth functions to get,

\begin{align*}
 \mathbb{E}[(\ell(w;(x,y)))^2]
      & \leq 2 \mathbb{E}[\norm{\nabla \ell(w;(x,y))}^2\gamma^2\log{4/\beta}   + 2\ybound^2 \\
    &\leq  8 \tilde H\mathbb{E}[\ell(w;(x,y))]\gamma^2\log{4/\beta} + 2\ybound^2 \\
    & = 8\tilde HL(w;\cD)\gamma^2\log{4/\beta} + 2\ybound^2 
\end{align*}

Plugging the above in Eqn \eqref{eqn:boosting_bernstein} gives us,

\begin{align}
\label{eqn:boosting-eqn}
\nonumber
    \abs{L(w;\cD)-\hat L(w;S)}
    &\leq \frac{\lbound \log{2/\beta}}{n} +
    4\sqrt{\frac{ \br{\tilde HL(w;\cD)\gamma^2\log{4/\beta} + \ybound^2} \log{1/\beta}}{n}}\\
    &\leq \frac{\lbound \log{2/\beta}}{n} +
    4\sqrt{\frac{\tilde H L(w;\cD)}{n}}\gamma \log{4/\beta} + \frac{\ybound\sqrt{\log{2/\beta}}}{\sqrt{n}}.
\end{align}

A simple application of AM-GM inequality gives,
\begin{align*}
    \hat L(w;S) \leq \br{1+\frac{4\gamma \log{4/\beta}\sqrt{\tilde H}}{\ybound\sqrt{n}}}L(w;\cD) + \frac{4\gamma\log{4/\beta}\ybound\sqrt{\tilde H}}{\sqrt{n}}+ \frac{\ybound\sqrt{\log{2/\beta}}}{\sqrt{n}}
\end{align*}

This proves the first part of the lemma. For the second part, we  use the following fact about non-negative numbers $A, B, C$ \cite{bousquet2003introduction} (see after proof of Theorem 7)
\begin{align*}
    A \leq B+ C\sqrt{A} \implies A \leq B + C^2 + \sqrt{B}C
\end{align*}

Thus, from Eqn. \eqref{eqn:boosting-eqn},
\begin{align*}
     L(w;\cD) &\leq \hat L(w;S) +\frac{\lbound \log{2/\beta}}{n}+ \frac{\ybound\sqrt{\log{2/\beta}}}{\sqrt{n}} + \frac{16\tilde H\gamma^2\operatorname{log}^2(4/\beta)}{n} \\
     &+ \frac{4\gamma\log{4/\beta}\sqrt{\tilde H}}{\sqrt{n}} \br{\sqrt{\hat L(w;S)} +\sqrt{\frac{\lbound \log{2/\beta}}{n}}+ \sqrt{\frac{\ybound\sqrt{\log{2/\beta}}}{\sqrt{n}}}} \\
     & \leq \hat L(w;S) + \frac{4\gamma\log{4/\beta}\sqrt{\tilde H\hat L(w;S)}}{\sqrt{n}}+ \frac{\lbound \log{2/\beta}}{n}+ \frac{\ybound\sqrt{\log{2/\beta}}}{\sqrt{n}} \\
     &+ \frac{16\tilde H\gamma^2\operatorname{log}^2(4/\beta)}{n} 
     + \frac{4\gamma\sqrt{\tilde H\lbound }\operatorname{log}^{3/2}(4/\beta)}{n}+ \frac{4\gamma\sqrt{\tilde H\ybound}\operatorname{log}^{5/4}(2/\beta)}{n^{3/4}} \\
     & \leq  \br{1+\frac{4\gamma \log{4/\beta}\sqrt{\tilde H}}{\ybound\sqrt{n}}}\hat L(w;S) + \frac{4\gamma\log{4/\beta}\ybound\sqrt{\tilde H}}{\sqrt{n}}+ \frac{\lbound \log{2/\beta}}{n}\\
     &+ \frac{\ybound\sqrt{\log{2/\beta}}}{\sqrt{n}} 
     + \frac{16\tilde H\gamma^2\operatorname{log}^2(4/\beta)}{n} +
     \frac{4\gamma\sqrt{\tilde H\lbound }\operatorname{log}^{3/2}(4/\beta)}{n} \\&+ \frac{4\gamma\sqrt{\tilde H\ybound}\operatorname{log}^{5/4}(2/\beta)}{n^{3/4}}
\end{align*}
where the last inequality follows from AM-GM inequality. Simplifying the expressions yields the claimed bound.
\end{proof}

\begin{proof}[Proof of Theorem \ref{thm:boosting}]

Since the models $\bc{\hat w_i}_{i=1}^m$ are trained on disjoint datasets, by parallel composition $\bc{\hat w_i}_{i=1}^m$ satisfies $\br{\frac{\epsilon}{2},\frac{\delta}{2}}$-DP. 
We know that probability at least $1-\frac{\delta}{2}$, $\ell(w;(x,y))\leq 
\lbounddp$. Thus conditioning on this event, from the guarantee of the report noisy max procedure, we have that it satisfies $\br{\frac{\epsilon}{2}}$-DP. 
The privacy proof thus follows from absorbing the failure probability into $\delta$ part and adaptive composition.

We proceed to the utility part.
Let $\tilde w$ be the model among $\bc{\hat w_i}_{i=1}^m$ with minimum empirical error on the set $S_{m+1}$. The excess risk of each $\hat w_i$ is bounded as,
\begin{align*}
    \mathbb{E}[L(\hat w_i;\cD)]  - L(w^*;\cD)\leq \err\br{\frac{n}{m+1},\frac{\epsilon}{2},\gamma}
\end{align*}
From Markov's inequality, with probability at least $3/4$, $L(\hat w_i;\cD) \leq L(w^*;\cD) + 4\err\br{\frac{n}{m+1},\frac{\epsilon}{2}}$. 
From independence of $\bc{w_i}_{i=1}^m$, with probability at least $1-1/4^m=1-\frac{\beta}{4}$, there exists one model, say $\hat w_{i^*}$, such $L(\hat w_{i^*};\cD) \leq L(w^*;\cD) + 4\err\br{\frac{n}{m+1},\frac{\epsilon}{2}}$.

Also, from the guarantee of Report-Noisy-Max, we have that with probability at least $1-\beta/4$
\begin{align*}
    L(\hat w;S_{m+1}) \leq L(\tilde w;S_{m+1}) + \frac{\Delta(\gamma,\beta/4)(m+1)\operatorname{log}^2{(4m/\delta)}}{n\epsilon}
\end{align*}

Now, we apply Lemma \ref{lem:boosting-concentration}. From a union bound, with probability at least $1-\frac{\beta}{2}$, all $\bc{w_i}_{i=1}^m$ satisfy the inequalities in Lemma \ref{lem:boosting-concentration} with $\beta$ substituted as $\frac{\beta}{2m}$.

Thus, for the output $\hat w$, probability at least $1-\frac{\beta}{2}$,

\begin{align*}
     &L(\hat w;\cD) \\&\leq \br{1+T\br{\frac{n}{m+1},\frac{\beta}{2m}}}L(\hat w;S_{m+1})+V\br{\frac{n}{(m+1)},\frac{\beta}{2m}} \\
     & \leq\br{1+T\br{\frac{n}{m+1},\frac{\beta}{2m}}}L(\tilde w;S_{m+1})+  \br{1+T\br{\frac{n}{m+1},\frac{\beta}{2m}}}\frac{\Delta(\gamma,\beta/4)(m+1)\operatorname{log}^2{(4m/\delta)}}{n\epsilon}\\&+ V\br{\frac{n}{(m+1)},\frac{\beta}{2m}} \\
      & \leq\br{1+T\br{\frac{n}{m+1},\frac{\beta}{2m}}}L(w_{i^*};S_{m+1})+  \frac{2\Delta(\gamma,\beta/4)(m+1)\operatorname{log}^2{(4m/\delta)}}{n\epsilon}+ V\br{\frac{n}{(m+1)},\frac{\beta}{2m}} 
\end{align*}

where in the above we use that $T\br{\frac{n}{m+1}} \leq 1$ given the lower bound on $n$ and the setting of $m$.
Furthermore, the last inequality follows because $\tilde w$ has lowest empirical risk on $S_{m+1}$.
Let $W(n,m,\beta) = \frac{2\Delta(\gamma,\beta/4)(m+1)\operatorname{log}^2{(4m/\delta)}}{n\epsilon}+ V\br{\frac{n}{(m+1)},\frac{\beta}{2m}}$.
We now apply the other guarantee in Lemma \ref{lem:boosting-concentration} and the fact that $w_{i^*}$ has small excess risk. With probability at least $1-\delta/2$,
\begin{align*}
     &L(\hat w;\cD)\\ &\leq\br{1+T\br{\frac{n}{m+1},\frac{\beta}{2m}}}^2L(w_{i^*};\cD) +\br{1+ T\br{\frac{n}{m+1},\frac{\beta}{2m}}}U\br{\frac{n}{m+1},\frac{\beta}{2m}} + W(n,m,\beta) \\
     & \leq L(w^*;\cD)+ 2T\br{\frac{n}{m+1},\frac{\beta}{2m}}L(w^*;\cD) + 4\err\br{\frac{n}{m+1},\frac{\epsilon}{2}} \\&+ 8T\br{\frac{n}{m+1},\frac{\beta}{2m}}4\err\br{\frac{n}{m+1},\frac{\epsilon}{2}} 
      + 2U\br{\frac{n}{m+1},\frac{\beta}{2m}} + W(n,m,\beta)
\end{align*}

Let $X(n,m,\beta)=4\err\br{\frac{n}{m+1},\frac{\epsilon}{2}} + 8T\br{\frac{n}{m+1},\frac{\beta}{2m}}4\err\br{\frac{n}{m+1},\frac{\epsilon}{2}} 
+ 2U\br{\frac{n}{m+1},\frac{\beta}{2m}} + W(n,m,\beta)$.
Note that $m=4\log{4/\beta}$ and $T\br{\frac{n}{m+1},\frac{\beta}{2m}} \leq \frac{16\gamma\operatorname{log}^4\br{2/\beta}\sqrt{H}}{\norm{\cY}\sqrt{n}}$. Substituting this and 
using the fact that, $L(w^*;\cD)\leq L(0;\cD) \leq \ybound^2$, we get that with probability at least $1-\delta$,
\begin{align*}
     L(\hat w;\cD) & \leq L(w^*;\cD) + \frac{16\gamma\operatorname{log}^4\br{2/\beta} \sqrt{H}\ybound}{\sqrt{n}} + X(n,4\log{2/\beta},\beta)
\end{align*}

 Substituting and simplifying the $X(n,4\log{4/\beta},\beta)$ we have that 
\begin{align*}
    &X(n,4\log{2/\beta},\beta) \\&\leq 12\err\br{\frac{n}{4\log{4/\beta}},\frac{\epsilon}{2},\gamma} + \frac{2\Delta(\gamma,\beta/4)\operatorname{log}^3{(4/\beta)}\operatorname{log}{(2/\delta)}}{n\epsilon} +\frac{2\lboundtwo \operatorname{log}^4\br{8/\beta}}{n}\\&+ \frac{16\gamma \operatorname{log}^4\br{8/\beta}\sqrt{\tilde H}\ybound}{\sqrt{n}} +  \frac{16\ybound\operatorname{log}^4(8/\beta)}{\sqrt{n}} +\frac{128\tilde H\gamma^2\operatorname{log}^4(8/\beta)}{n}
\end{align*}
Hence, with probability at least $1-\beta$,
\begin{align*}
 &L(\hat w;\cD) \\& \leq L(w^*;\cD)+
    12\err\br{\frac{n}{4\log{4/\beta}},\frac{\epsilon}{2},\gamma} +
 \frac{2\Delta(\gamma,\beta/4)\operatorname{log}^3{(4/\beta)}\operatorname{log}{(2/\delta)}}{n\epsilon}  +\frac{2\lboundtwo \operatorname{log}^4\br{8/\beta}}{n}\\&+ \frac{32\gamma \operatorname{log}^4\br{8/\beta}\sqrt{\tilde H}\ybound}{\sqrt{n}} +  \frac{16\ybound\operatorname{log}^4(8/\beta)}{\sqrt{n}} +\frac{128\tilde H\gamma^2\operatorname{log}^4(8/\beta)}{n}
\end{align*}
\end{proof}

\subsection{Boosting the JL Method}
\begin{theorem}
The boosting procedure (Algorithm \ref{alg:boosting}) using the JL method (Algorithm \ref{alg:jl-constrained-dp-erm}) as Algorithm $\cA$ satisfies $(\epsilon, \delta)$-DP, and with probability at least $1-\beta$, its output $\hat w_{i^*}$ has excess risk,
\begin{align*}
    L(\hat w_{i^*};\cD) - L(w^*;\cD) &\leq \tilde O\Bigg(\frac{\sqrt{H}B\norm{\cX}\norm{\cY}}{\sqrt{n}} +\frac{\br{\br{\sqrt{H}B\norm{\cX}}^{4/3}\norm{\cY}^{2/3}+ \br{\sqrt{H}B\norm{\cX}}^2}}{(n\epsilon)^{2/3}} \\&+ \frac{\br{\ybound^2 + HB^2\xbound^2}}{n\epsilon} + \frac{\br{\ybound+\sqrt{ H}B\xbound}}{\sqrt{n}}
    \Bigg)
\end{align*}

\end{theorem}
\begin{proof}
For the JL method, we consider the boosting procedure in the $k$ dimension space -- this is only for the sake of analysis and the algorithm remains the same.
In particular, consider the distribution to $\Phi\cD$ which, when sampled from, gives the data point $\br{\Phi x, y}$ where $(x,y)\sim \cD$.

Suppose the boosting procedure gives the following $k$ dimensional models: $\tilde w_1, \cdots \tilde w_{m}$; note that the norm of all these are bounded by $2B$. Let $\tilde w^* \in \arg\min_{\norm{w}\leq 2B}L(w;\Phi \cD)$. Since the outputs satisfy $\norm{\tilde w_i}\leq 2B$, the sub-Gaussian parameter $\gamma = O(B)$.
We now compute the other parameter $\Delta(\gamma, \beta)$, which is the high probability bound on loss. Note that for a fixed data point $(\Phi x,y)$, an $H$ smooth, non-negative, bounded at zero loss, at any point $w$ s.t. $\norm{w}\leq 2B$ is upper bounded by $ 2\br{\norm{\cY}^2 + 4B^2H\norm{\Phi x}^2}$. 
From the JL guarantee, with the given $k$, the term $\norm{\Phi x}^2 \leq 2\norm{\cX}$ with probability at least $1-\beta/4$.
This gives us $\Delta(2B,\beta) = 2\br{\norm{\cY}^2 + 16\norm{\cX}^2B^2}$.

We now invoke \ref{thm:boosting} substituting $\Delta$, $\gamma$ and $\tilde H = H\xbound^2$ to get
that with probability at least $1-\frac{\beta}{2}$ output satisfies,
\begin{align*}
   L(\tilde w_{i^*};\Phi\cD) & \leq L(\tilde w^*;\Phi\cD) + \frac{128B\sqrt{H}\xbound\ybound 
    \operatorname{log}^4{(8/\beta)}}{\sqrt{n}} + \alpha\br{\frac{n}{4\log{(8/\beta)}},\frac{\epsilon}{2},2B}\\&+ \frac{128\br{\ybound^2 + HB^2\xbound^2} \operatorname{log}^4{(8/\beta)}}{n\epsilon} + \frac{128\br{\ybound+\sqrt{ H}B\xbound} \operatorname{log}^4{(8/\beta)}}{\sqrt{n}}
\end{align*}

Define $W:=\frac{128B\sqrt{H}\xbound\ybound
    \operatorname{log}^4{(8/\beta)}}{\sqrt{n}}+ \frac{128\br{\ybound^2 + HB^2\xbound^2} \operatorname{log}^4{(8/\beta)}}{n\epsilon} + \frac{128\br{\ybound+\sqrt{ H}B\xbound} \operatorname{log}^4{(8/\beta)}}{\sqrt{n}}$.
The excess risk of the final output $\hat w_{i^*} = \Phi^\top \tilde w_{i^*}$ is bounded as,
\begin{align*}
    L(\hat w_{i^*};\cD) - L(w^*;\cD) &=   L(\tilde w_{i^*};\Phi\cD) -  L(w^*;\cD) \\
    & \leq L(\tilde w^*;\Phi\cD) + \alpha\br{\frac{n}{4\log{8/\beta}},\frac{\epsilon}{2},2B} + W - L(w^*;\cD) \\
    &\leq L(\Phi w^*;\Phi\cD)+ \alpha\br{\frac{n}{4\log{8/\beta}},\frac{\epsilon}{2},2B} + W - L(w^*;\cD) \\
    & \leq \alpha\br{\frac{n}{4\log{8/\beta}},\frac{\epsilon}{2},2B} + W + \frac{H}{2}\abs{\ip{\Phi x}{\Phi w^*} - \ip{x}{w^*}}^2
\end{align*}
where the last inequality follows from smoothness and that $\nabla L(w^*;\cD)=0$. Finally, from the JL property, with probability at least $1-\frac{\beta}{2}$, $\abs{\ip{\Phi x}{\Phi w^*} - \ip{x}{w^*}}^2 \leq \alpha^2\norm{w^*}^2\xbound^2$. Combining and substituting the values of $\alpha$ and $\alpha\br{\frac{n}{4\log{8/\beta}},\frac{\epsilon}{2}}$ from \cref{thm:smooth-jl} gives the claimed result.
\end{proof}
\subsection{Boosting Output Perturbation Method}

\begin{theorem} \label{thm:boosted-smooth-output-perturbation}
The boosting procedure (Algorithm \ref{alg:boosting}) using the output perturbation method (Algorithm \ref{alg:constrained-reg-erm-output-perturbation}) as input Algorithm $\cA$ satisfies $(\epsilon, \delta)$-DP, and with probability at least $1-\beta$, its output $\hat w_{i^*}$ has excess risk,
\begin{align*}
    L(\hat w_{i^*};\cD) - L(w^*;\cD) &\leq \tilde O\Bigg(\frac{\sqrt{H}B\norm{\cX}\norm{\cY} + \norm{\cY}^2}{\sqrt{n}} +\frac{\br{\br{\sqrt{H}B\norm{\cX}}^{4/3}\norm{\cY}^{2/3}+ \br{\sqrt{H}B\norm{\cX}}^2}}{(n\epsilon)^{2/3}} \\&+ \frac{\br{\ybound^2 + HB^2\xbound^2}}{n\epsilon} + \frac{\br{\ybound+\sqrt{ H}B\xbound}}{\sqrt{n}}
    \Bigg)
\end{align*}
\end{theorem}
\begin{proof}
Firstly, note that $\tilde H = H\xbound^2$.
We now compute $\gamma$ and $\Delta$ to invoke \cref{thm:boosting}. Since the algorithm simply adds Gaussian noise of variance $\sigma^2\bbI_d$ to a vector $\tilde w$ where $\norm{\tilde w}\leq B$, we have that $\gamma^2 \leq B^2 + \sigma^2$. For the bound on loss parameter $\Delta$, by direct computation, $\Delta(\gamma, \beta) \leq 2\br{\norm{\cY}^2 + H\ip{\tilde w + \xi}{x}^2} \leq 2\norm{\cY}^2 + 2HB^2\norm{\cX}^2 + 2H\ip{\xi}{x}^2 \leq 2\norm{\cY}^2 + 2HB^2\norm{\cX}^2 + 2H\sigma^2\log{1/\beta}$ where the last inequality follows since $\ip{\xi}{x} \sim \cN(0,\norm{x}^2)$. Plugging these and the value of $\sigma^2$ from Theorem \ref{thm:smooth-output-perturbation} into Theorem \ref{thm:boosting} gives the claimed bound.
\end{proof}

\section{Non-private Lower Bounds} \label{app:nonpriv-lower}
We first note a simple one-dimensional lower bound.
\begin{theorem}
(Implicit in \cite{SST10}) Let $\xbound \geq 1$ and $H\geq 2$. For any Algorithm $\cA$, there exists a 1-dimensional $H$-smooth non-negative GLM, $\ell:\re\times(\cX\times\cY) \mapsto \re$, and a distribution $\cD$ over $(\cX\times\cY)$ with $\mnorm = \Theta(\min\bc{\ybound,B}$ such that the excess population risk of the output of $\cA$ when run on $S\sim\cD^n$ is lower bounded as $\Omega\br{\frac{\min\bc{\ybound,B\xbound}}{\sqrt{n}}}$.
\end{theorem}
We remark that this requires a slight modification of the example used in \cite{SST10}. Specifically, therein the loss function is defined as 
\begin{align*}
    \ell(w,(x,y)) = \begin{cases}
    (w-y)^2 & |w-y| \leq \frac{1}{2} \\
    |w-y| - 1/4 & |w-y| \geq \frac{1}{2}
    \end{cases}
\end{align*}
with $y \in \bc{\pm 1}$ and $x=1$.
Our statement is obtained by setting the domain of labels as $\bc{\pm\min\bc{B,\ybound}}$. 
A reduction in \cite{shamir_2015} enables lower bounds from problem instances with general $\xbound$ and $\mnorm = R$ to instances with $\xbound=1$ to $\mnorm = R\xbound$.

We now show that the lower bounds presented in \cite{shamir_2015} to the unconstrained setting. 
We start by stating the original bound from \cite{shamir_2015} which holds for squared loss. 
\begin{theorem}\label{thm:shamir}
Let $\ell(w,(x,y)) = \frac{1}{2}(\ip{w}{x} - y)^2$ be the squared loss function and $B>0$. Then for any algorithm, $\cA$, there exists a distribution $\cD$ over $(\cX\times\cY)$ and a constant $C$ such that for $S\sim\cD^n$ it holds that
$\ex{}{L(\cA(S);\cD) - L(w_B^*;\cD)} \geq C\min\bc{\ybound^2, \frac{B^2\xbound^2+d\ybound^2}{n},\frac{B\ybound\xbound}{\sqrt{n}}}$,
where $w_B^* = \argmin\limits_{w\in\ball{B}}\bc{L(w;\cD)}$.
\end{theorem}
We now make three observations.
First we note that this Theorem holds even for $\cA(S)\notin \ball{B}$. Second we note that the construction of the distribution in the above Theorem is such that $\min\limits_{w^*\in\ball{B}}\bc{L(w;\cD)} = \min\limits_{w^*\in\re^d}\bc{L(w;\cD)}$. Finally, note that $\frac{H}{2}(\ip{w}{x}-\frac{y}{\sqrt{H}})^2 = \frac{1}{2}(\sqrt{H}\ip{w}{x}-y)^2$. This gives the following corollary. 
\begin{corollary}
Let $B>0$. Then for any algorithm, $\cA$, there exists a distribution $\cD$ over $(\cX\times\cY)$  and an $H$-smooth non-negative GLM, $\ell:\re^d \times (\cX\times\cY) \mapsto \re$, 
with minimizer 
$w^* = \argmin\limits_{w\in\re^d}\bc{L(w;\cD)}$ such that $\mnorm=B$
and for $S\sim\cD^n$ it holds that
$$\ex{}{L(\cA(S);\cD) - L(w^*;\cD)} = \Omega\bc{\min\bc{\ybound^2, \frac{HB^2\xbound^2+d\ybound^2}{n},\frac{\sqrt{H}B\ybound\xbound}{\sqrt{n}}}}.$$
\end{corollary}

\section{Additional Results}
\begin{lemma}
\label{lem:jl}
Let $\Phi \in \mathbb{R}^{d\times k}$ be a random matrix such that for any $u \in \mathbb{R}^d$, with probability at least $1-\beta$, $(1-\alpha)\norm{u}^2 \leq \norm{\Phi u}^2  \leq (1+\alpha)\norm{u}^{2}$. Then for any $u, v$, with probability at least $1-2\beta$, $\abs{\ip{\Phi u}{\Phi v} - \ip{u}{v}}\leq \alpha \norm{u}\norm{v}$.
\end{lemma}
\begin{proof}
Firstly, note that it suffices to prove the result for unit vectors $u$ and $v$. 
From the norm preservation result, with probability at least $1-2\beta$, we have that,
\begin{align*}
    (1-\alpha)\norm{u+v}^2 &\leq \norm{\Phi(u+v)} \leq (1+\alpha)\norm{u+v}^2  \\
    (1-\alpha)\norm{u-v}^2 &\leq \norm{\Phi(u-v)} \leq (1+\alpha)\norm{u-v}^2  \\
\end{align*}
Therefore, we have
\begin{align*}
    \ip{\Phi u}{\Phi v} &= \frac{1}{4}\br{\norm{\Phi\br{u+v}}^2 - \norm{\Phi\br{u-v}}^2} \\
    &\leq \frac{1}{4}\br{\br{1+\alpha}\norm{u+v}^2 - (1-\alpha)\norm{u-v}^2} \\
    & \leq \ip{u}{v} + \alpha
\end{align*}
This gives us that $\ip{\Phi u}{\Phi v} \leq \ip{u}{v}+\alpha$. The other inequality follows in the same way.
\end{proof}

\end{document}